\def\aaaianonymous{False}
\documentclass[letterpaper]{article} 

    \usepackage{aaai2026}              

\usepackage{times}  
\usepackage{helvet}  
\usepackage{courier}  
\usepackage[hyphens]{url}  
\usepackage{graphicx} 
\usepackage{natbib}  
\usepackage{caption} 
\usepackage{algorithm}
\usepackage{algorithmic}

\usepackage{newfloat}
\usepackage{listings}

\usepackage{subcaption}
\usepackage{booktabs}
\usepackage{array}
\usepackage{makecell}
\usepackage{amsmath}
\usepackage{amssymb}
\usepackage{amsfonts}       

\DeclareCaptionStyle{ruled}{labelfont=normalfont,labelsep=colon,strut=off} 
\floatstyle{ruled}
\newfloat{listing}{tb}{lst}{}
\floatname{listing}{Listing}

\ifdefined\aaaianonymous
    \title{W2S-AlignTree: Weak-to-Strong Inference-Time Alignment for \\Large Language Models via Monte Carlo Tree Search}
\else
    \title{AAAI Press Formatting Instructions \\for Authors Using \LaTeX{} --- A Guide}
\fi

\author{
	Zhenyu Ding\textsuperscript{\rm 1},
	Yuhao Wang\textsuperscript{\rm 2},
	Tengyue Xiao\textsuperscript{\rm 1},
	Haoying Wang\textsuperscript{\rm 1},
	Caigui Jiang\textsuperscript{\rm 1},
	Ning Ding\textsuperscript{\rm 1} \thanks{Corresponding author.}
}
\affiliations{
    \textsuperscript{\rm 1}State Key Laboratory of Human-Machine Hybrid Augmented Intelligence, National Engineering Research Center for Visual Information and Applications, and Institute of Artificial Intelligence and Robotics, Xi'an Jiaotong University, Xi'an, China\\
    \textsuperscript{\rm 2}School of Computer Science and Technology, Xi'an Jiaotong University, Xi'an, China
}

\begin{document}

\maketitle

\begin{abstract}
Large Language Models (LLMs) demonstrate impressive capabilities, yet their outputs often suffer from misalignment with human preferences due to the inadequacy of weak supervision and a lack of fine-grained control. Training-time alignment methods like Reinforcement Learning from Human Feedback (RLHF) face prohibitive costs in expert supervision and inherent scalability limitations, offering limited dynamic control during inference.
Consequently, there is an urgent need for scalable and adaptable alignment mechanisms.
To address this, we propose W2S-AlignTree, a pioneering plug-and-play inference-time alignment framework that synergistically combines Monte Carlo Tree Search (MCTS) with the Weak-to-Strong Generalization paradigm for the first time.
W2S-AlignTree formulates LLM alignment as an optimal heuristic search problem within a generative search tree.
By leveraging weak model's real-time, step-level signals as alignment proxies and introducing an Entropy-Aware exploration mechanism, W2S-AlignTree enables fine-grained guidance during strong model's generation without modifying its parameters. 
The approach dynamically balances exploration and exploitation in high-dimensional generation search trees. 
Experiments across controlled sentiment generation, summarization, and instruction-following show that W2S-AlignTree consistently outperforms strong baselines. 
Notably, W2S-AlignTree raises the performance of \texttt{Llama3-8B} from $1.89$ to $2.19$, a relative improvement of $15.9\%$ on the summarization task.
\end{abstract}

\ifdefined\aaaianonymous
\begin{links}
    \link{Code}{https://github.com/alexzdy/W2S-AlignTree}
\end{links}
\else
\begin{links}
    \link{Code}{https://aaai.org/example/code}
    \link{Datasets}{https://aaai.org/example/datasets}
    \link{Extended version}{https://aaai.org/example/extended-version}
\end{links}
\fi

\ifdefined\aaaianonymous
\section{Introduction}

\begin{figure}[!ht]
  \centering
  \includegraphics[width=0.898\linewidth]{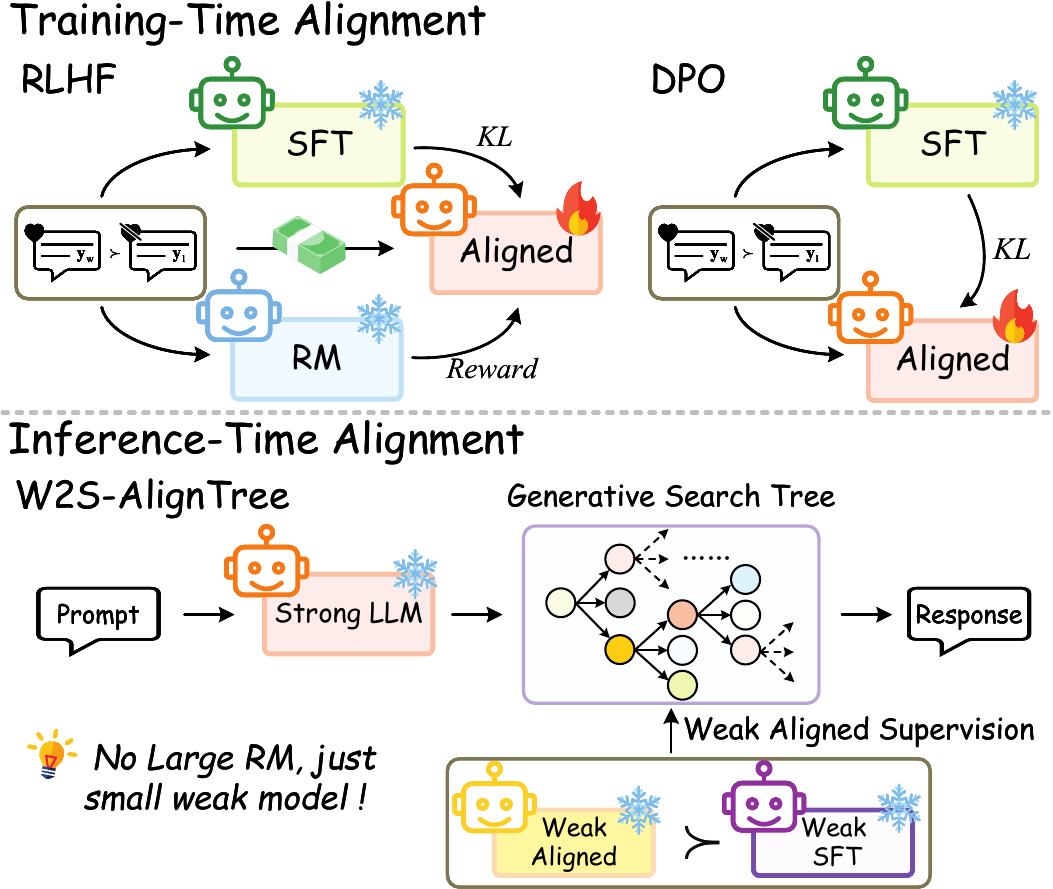}
  \caption{W2S-AlignTree vs. Training-Time Alignment.
Unlike RLHF and DPO, W2S-AlignTree enables fine-grained inference-time alignment using weak model signals—without costly reward models or parameter updates.}
  \label{fig:teaser}
\end{figure}

Large Language Models (LLMs) have demonstrated capabilities in natural language understanding, text generation, and complex reasoning that approach or even surpass human performance. 
However, a crucial issue is the misalignment between their behaviors and human values, frequently manifesting as biased, harmful, or false content~\cite{weidinger2022taxonomy}. 
Both academia and industry have increasingly focused on developing LLM alignment methods to ensure helpful, honest, and harmless responses.
The prevailing alignment paradigm is Reinforcement Learning from Human Feedback (RLHF), which typically involves supervised fine-tuning (SFT), reward model training~\cite{christiano2017deep} and Proximal Policy Optimization (PPO)~\cite{ouyang2022training}.
As shown in Fig.~\ref{fig:teaser}, despite its empirical successes, RLHF faces several critical challenges.
First, RLHF relies heavily on large-scale, high-quality human-annotated data to train reward models or reinforcement learning algorithms, which are known to be unstable and computationally expensive.
Parameter‑efficient fine‑tuning methods like LoRA~\cite{hu2022lora} mitigate compute costs by freezing base model weights, but they consequently constrain generative flexibility. 
Additionally, Direct Preference Optimization (DPO)~\cite{rafailov2023direct} and its variants~\cite{zhou2023beyond,meng2024simpo} reframe preference learning as contrastive loss minimization, eliminating explicit reward modeling and RL sampling to improve training stability and efficiency.
However, DPO and RLHF both face the same problem: they rely on sequence‑level and post‑hoc feedback available only during training, which leaves them unable to provide immediate, fine‑grained control at inference time.
More fundamentally, as LLMs grow in scale, their behaviors may exceed the capacity of human annotation or other limited supervision, making feedback signals inadequate for aligning them~\cite{openai2023introducing}.
Early attempts like OpenAI’s Weak‑to‑Strong Generalization (W2SG)~\cite{burns2023weak} seek to use weaker supervision to align stronger models, yet remain largely preliminary and offer no real-time control for inference.

With the emergence of frontier LLMs~\cite{o1,guo2025deepseek}, Inference-time Scaling has become a promising paradigm.
It enhances model performance by strategically allocating additional computing resources during the model inference phase, offering a new avenue to overcome the optimization bottleneck in the training phase. 
Methods such as Chain-of-Thought (CoT)~\cite{wei2022chain} and Tree-of-Thought (ToT)~\cite{yao2023tree} have significantly enhanced LLM performance on complex reasoning by guiding multi-step or parallel thinking.
Building on this trend, inference-time alignment methods, including CBS~\cite{zhou2024weak} and TPO~\cite{li2025test}, have also appeared. While these methods guide model outputs via preference signals without parameter updates, they often struggle to explore complex generative spaces or enable fine-grained control.

Monte Carlo Tree Search (MCTS)~\cite{silver2016mastering, silver2017mastering} addresses these limitations by employing the Upper Confidence bounds applied to Trees (UCT)~\cite{UCT} to balance node visits and returns in large search spaces, offering a promising approach to improve LLM inference.
While MCTS has already been successfully applied to enhance mathematical reasoning~\cite{zhang2024accessing, qi2024mutual} and task planning~\cite{zhang2024rest, zhai2025enhancing}, its potential for inference-time alignment remains largely unexplored. 
Even in alignment-related work MCTS-DPO~\cite{xie2024monte}, MCTS is primarily used for offline data generation rather than real-time model guidance. 
Critically, within the W2SG framework, leveraging MCTS to dynamically align strong models under weak supervision represents an underexplored key issue for future research.

To bridge the gap between powerful LLMs and effective alignment under weak supervision, this paper proposes \textbf{W2S-AlignTree}, the first framework that integrates MCTS with the W2SG paradigm.
As indeicated in Fig.~\ref{fig:teaser}, we formulate preference alignment as a search process over a generative search tree, where a weak model provides dynamic guidance to efficiently and scalably explore the strong model’s response space.

The main contributions of this paper are as follows:
\begin{itemize}
    \item We introduce \textbf{W2S-AlignTree}, a plug-and-play MCTS-based alignment framework built on a ``weak guidance-strong exploration" mechanism. 
    By injecting preference proxy signals from a weak LLM, it enables dynamic and fine-grained control over strong LLMs with unified step-level guidance and sequence-level evaluation.
    \item We design an Entropy-Aware Prioritized UCT (EA-PUCT) selection rule that integrates policy entropy and prior probabilities to adaptively capture uncertainty, reducing premature convergence and improving trajectory diversity and quality in complex alignment tasks.
    \item Comprehensive experiments show W2S-AlignTree significantly boosts LLM alignment on a broad spectrum of challenging tasks, including controlled-sentiment generation, summarization, and instruction-following.
\end{itemize}

\section{Preliminaries \& Problem}
\subsection{Preliminaries}
\paragraph{RLHF.}
Assume the input prompt $\mathbf{x}$ from probability distribution $p(\mathbf{x})$, and a complete model response $\mathbf{y}$. 
The model to be aligned is denoted as $\pi(\mathbf{y}|\mathbf{x})$, while the reference model is denoted as $\pi_{\text{ref}}(\mathbf{y}|\mathbf{x})$ (e.g., SFT model). A reward function $r(\mathbf{x}, \mathbf{y})$  measures the quality of the responses. The objective of RLHF can be written as:
\begin{equation}
\begin{aligned}
\arg\max_{\pi}\quad
& \mathbb{E}_{\mathbf{x}\sim p(\mathbf{x}),\,\mathbf{y}\sim\pi(\mathbf{y}|\mathbf{x})}\bigl[r(\mathbf{x},\mathbf{y})\bigr] \\
\text{s.t.}\quad
& \mathbb{E}_{\mathbf{x}\sim p(\mathbf{x})}\Bigl[\mathbb{D}_{\mathrm{KL}}\bigl(\pi(\mathbf{y}|\mathbf{x})\bigm\|\pi_{\text{ref}}(\mathbf{y}|\mathbf{x})\bigr)\Bigr]\leq\epsilon,
\end{aligned}
\end{equation}
where $\mathbb{D}_{\mathrm{KL}}$ restricts the optimized model $\pi$ from deviating too much from the reference (unaligned) model $\pi_\text{ref}$.

For this constrained optimization, a globally optimal closed-form solution $\pi^*(\mathbf{y}|\mathbf{x})$ exists, and its relationship with the reward function can be expressed via the Lagrangian formulation~\cite{ouyang2022training}:
\begin{equation}
    r(\mathbf{x}, \mathbf{y}) = \beta \log \frac{\pi^*(\mathbf{y}|\mathbf{x})}{\pi_{\text{ref}}(\mathbf{y}|\mathbf{x})} + \beta \log Z(\mathbf{x}),
\end{equation}
where $Z(\mathbf{x})=\sum_{\mathbf{y}}\pi_{\text{ref}}(\mathbf{y}|\mathbf{x})\exp\big(\frac{1}{\beta}r(\mathbf{x},\mathbf{y})\big)$ denotes the partition function. This term acts as a normalization constant, ensuring that $\pi^*(\mathbf{y}|\mathbf{x})$ forms a valid distribution by summing to $1$ over all possible responses $\mathbf{y}$.
$\beta$ controls the KL regularization and implicitly scales the reward.

\paragraph{DPO.}
DPO~\cite{rafailov2023direct} streamlines RLHF by using its closed-form solution to cast the alignment objective as a Bradley–Terry contrastive learning problem~\cite{bradley1952rank}, thereby eliminating reward model training and RL sampling to improve differentiability and stability. 
Given a prompt $\mathbf{x}$ with corresponding accepted and rejected responses $\mathbf{y}_\text{w}$ and $\mathbf{y}_\text{l}$, it optimizes the objective about $\pi$:
\begin{equation}
\begin{aligned}
&\mathcal{L}_{\text{DPO}} \left( \pi; \pi_{\text{ref}} \right)
= -\mathbb{E}_{(\mathbf{x}, \mathbf{y}_\text{w}, \mathbf{y}_\text{l}) \sim \mathcal{D}} \\ 
&\left[ \log \sigma \left( \beta \log \frac{\pi(\mathbf{y}_\text{w}|\mathbf{x})}{\pi_{\text{ref}}(\mathbf{y}_\text{w}|\mathbf{x})} - \beta \log \frac{\pi(\mathbf{y}_\text{l}|\mathbf{x})}{\pi_{\text{ref}}(\mathbf{y}_\text{l}|\mathbf{x})} \right) \right],
\end{aligned}
\end{equation}
where $\mathcal{D}$ represents the preference dataset, and $\sigma(\cdot)$ is the sigmoid function.
DPO regards $\beta \log \frac{\pi(\mathbf{y}|\mathbf{x})}{\pi_{\text{ref}}(\mathbf{y}|\mathbf{x})}$ as an implicit reward, where $Z(\mathbf{x})$ is naturally canceled in the objective.

\subsection{Problem Formulation}
Despite the theoretical elegance and practical successes of RLHF and DPO, they both typically rely on sparse, sequence-level rewards, which are only available after full responses are generated.
This ``post-adjustment" paradigm hinders the provision of real-time feedback and fine-grained alignment during inference~\cite{rafailov2024r,shao2024deepseekmath}.
We address this by introducing a value function decomposition method, integrating the structural preference signals for real-time alignment guidance.

\paragraph{Definition 1 (Token-level Reward Decomposition).}
Consider LLM generation as a token-level Markov Decision Process (MDP): the state at $t$ is $s_t=(\mathbf{x},\mathbf{y}_{<t})$, and the action is the next token $a_t=y_t$ sampled from the vocabulary.
To obtain a dense reward, we decompose the sequence-level alignment reward $r(\mathbf{x}, \mathbf{y})$. 
Using the closed-form solution of $\pi^*(\mathbf{y}|\mathbf{x})$ and the chain rule, it can be expressed as:
\begin{equation}
\label{dense_reward}
    r(\mathbf{x},\mathbf{y})
    =\beta\log\frac{\pi^*(\mathbf{y}|\mathbf{x})}{\pi_{\text{ref}}(\mathbf{y}|\mathbf{x})}
    =\beta\sum_{t=1}^{|\mathbf{y}|}\log\frac{\pi^*(y_t|\mathbf{x},\mathbf{y}')}{\pi_{\text{ref}}(y_t|\mathbf{x},\mathbf{y}')},
\end{equation}
where $\mathbf{y}'=\mathbf{y}_{<t}$ represents the prefix up to token $t - 1$, $\mathbf{y}_{|\mathbf{y}|}$ is the \verb|EOS| token, and $Z(\mathbf{x})$ is omitted as it does not influence the objective.
This converts sparse sequence-level rewards into a continuous stream of token-level evaluations.

\paragraph{Definition 2 (Intermediate Value Function).}
We define an intermediate value function $V^*(\mathbf{x},\mathbf{y}')$ that represents the optimal expected future return for a partial sequence $\mathbf{y}'$ (see Appendix A.3 for details). 
By the Bellman optimality equations~\cite{puterman2014markov}, the cumulative log-likelihood ratio in Eq.~\ref{dense_reward} for $\mathbf{y}'$ relates to $V^*(\mathbf{x},\mathbf{y}')$:
\begin{equation}  
\label{eq:value or reward with y}
    \beta\log\frac{\pi^*(\mathbf{y}'|\mathbf{x})}{\pi_{\text{ref}}(\mathbf{y}'|\mathbf{x})} = 
\begin{cases} V^*(\mathbf{x},\mathbf{y}'), & \text{if } \mathbf{y}'\neq\mathbf{y} \\ 
r(\mathbf{x},\mathbf{y}), & \text{if } \mathbf{y}'=\mathbf{y}, \end{cases}
\end{equation}
where $V^*(\mathbf{x},\mathbf{y}')$ indicates the promise of a prefix for fully aligned responses, 
and the absence of rewards at intermediate steps matches sequence-level alignment's sparse-reward settings.
In other words, generating from  $(\mathbf{x},\mathbf{y}')$ with high $\beta\log\frac{\pi^*(\mathbf{y}'|\mathbf{x})}{\pi_{\text{ref}}(\mathbf{y}'|\mathbf{x})}$ has potential to yield a response $\mathbf{y}$ with high overall return $\beta\log\frac{\pi^*(\mathbf{y}|\mathbf{x})}{\pi_{\text{ref}}(\mathbf{y}|\mathbf{x})}$~\cite{zhou2024weak}.

\begin{figure*}[!ht]
  \centering
  \includegraphics[width=0.86\linewidth]{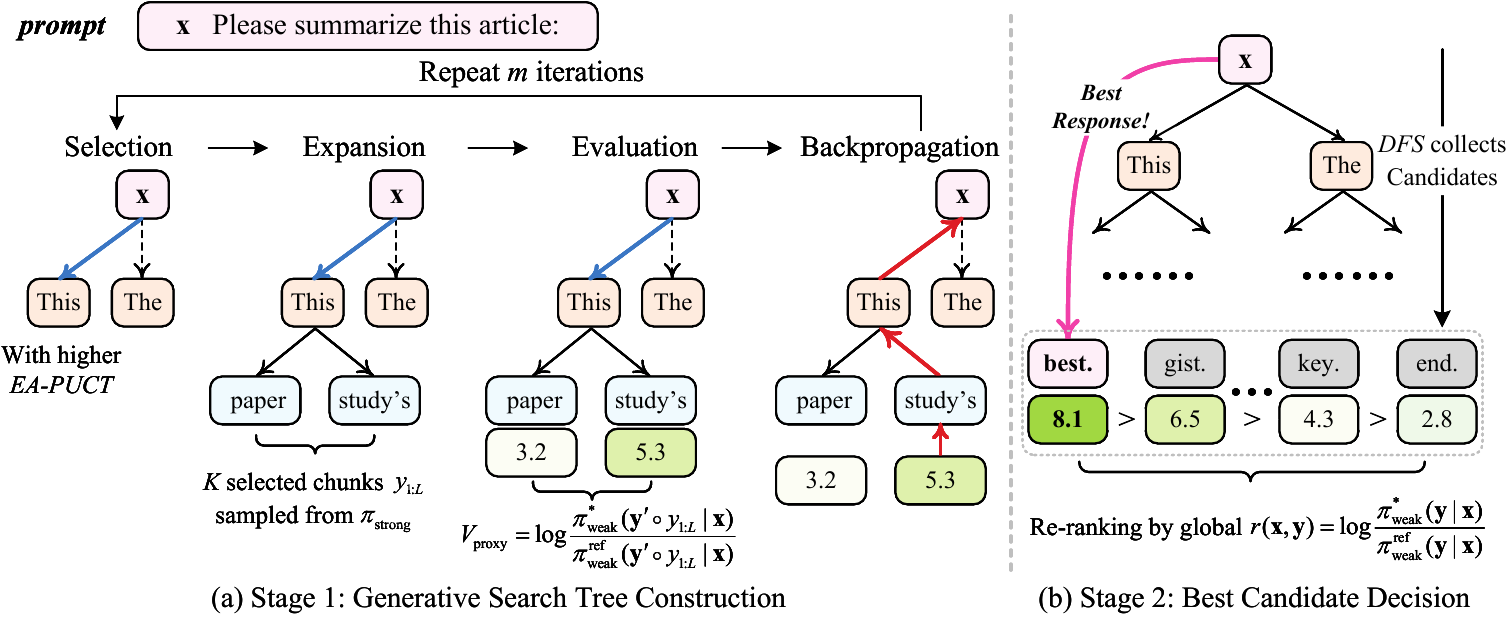}
  \caption{Dual-stage alignment process of W2S-AlignTree.
(a) MCTS constructs a generative search tree where candidate chunks are proposed by the strong model and evaluated with step-level proxy values from a weak model.
(b) Among all explored paths, W2S-AlignTree decides the response by globally re-ranking based on complete sequence-level alignment scores.
}
  \label{fig:main}
\end{figure*}

\paragraph{Definition 3 (Weak-to-Strong Proxy Mapping).}

Considering the prohibitive cost of repeated training, we adopt the W2SG paradigm to achieve scalable alignment at inference time. 
Our core idea is a proxy mapping that enables an unaligned strong model $\pi_{\text{strong}}$ to ``steal" preferences from a pre-aligned weak model $\pi_{\text{weak}}^{*}$. 
During the strong model's inference process, we utilize a prefix-dependent proxy value function $V_{\text{proxy}}(\mathbf{x}, \mathbf{y}')$ based on Definition 2:
\begin{equation}
\label{eq:weak_reward_model}
V_{\text{proxy}}(\mathbf{x}, \mathbf{y}')  =  \log \frac{\pi_{\text{weak}}^{*}(\mathbf{y}'|\mathbf{x})}{\pi_{\text{weak}}^{\text{ref}}(\mathbf{y}'|\mathbf{x})},
\end{equation}
where $\beta$ is also ignored since it scales all paths equally.
This $V_{\text{proxy}}$ provides dense, dynamic and step-by-step feedback rather than a static reward. 
This granular signal can deeply couple with search-based decoding process. 
At each generation step, $V_{\text{proxy}}$ directly modifies the strong model's sampling probabilities, steering its decoding to the weak model's preference. 
The precise integration of $V_{\text{proxy}}$ with search-based alignment will be detailed in the next section.

\section{Methodology}
This section introduces the proposed W2S-AlignTree, which formalizes LLM alignment as an optimal heuristic search problem during the inference phase. W2S-AlignTree uses a dual‑stage strategy: first, MCTS expands solution space steered by $V_{\text{proxy}}$; then, the globally optimal leaf node is selected as the final output. To adapt exploration to model uncertainty, we enhance UCT with an Entropy‑Aware bonus, allowing dynamic adjustment of exploration–exploitation based on the strong network's local entropy.
Please refer to Appendix A.5 for more details.

\subsection{Overview of W2S-AlignTree}

LLM generation requires strict adherence to specific alignment preferences, which can be precisely modeled as an optimal search problem over a \textbf{generative search tree}. 
In this tree, each node $s_t=(\mathbf{x}, \mathbf{y}')$ represents the MDP state with prompt $\mathbf{x}$ and current prefix $\mathbf{y}'$, and
each edge corresponds to the action $a_t=y_t$. 
A full root-to-leaf path forms a complete candidate response $\mathbf{y}$.
Given the exponential growth of the generation space, traditional greedy decoding often fails to find best solutions. 
To address this, W2S-AlignTree aims to steer the generation of the strong model $\pi_{\text{strong}}(y_t|\mathbf{x}, \mathbf{y}')$ at each step. 
This guidance signal originate not from a costly external reward model but from a proxy value $V_{\text{proxy}}$ derived from the weaker model as per Definition 3.
Our goal is to select the next token $y_t$ to maximize the following function:
\begin{equation}
\begin{split}
    &\quad\ \arg\max_{y_t} \mathcal{G}(\mathbf{x},\mathbf{y}',y_t) \\
    &= \arg\max_{y_t} [\log \pi_{\text{strong}}(y_t|\mathbf{x},\mathbf{y}') + V_{\text{proxy}}(\mathbf{x},\mathbf{y}' \circ y_t)],
\end{split}
\end{equation}
where $s' = (\mathbf{x},\mathbf{y}' \circ y_t)$ denotes the newly reached state.
The objective offers a principled unification of strong model's inherent generative capabilities with the alignment preferences of the weak model, making it naturally well-suited for the MCTS framework.
During the MCTS process, $V_{\text{proxy}}$ can be directly considered as the immediate reward $R(s')$ upon reaching the new state. 
To address the semantic discrepancy of $V_{\text{proxy}}$ when evaluating complete versus partial sequences due to Definition 2, W2S-AlignTree employs a dual-stage tree search to achieve inference-time alignment.

\subsection{Stage 1: Generative Search Tree Construction}
The first stage of W2S-AlignTree constructs a generative search tree. As demonstrated in Fig.~\ref{fig:main} (a), we incrementally approximate optimal solutions by four iterative phases:

\paragraph{Selection.}
The phase traverses the existing tree from the root node, repeatedly selecting the child node with the highest potential until an unvisited leaf node is reached for expansion. We employ an EA-PUCT selection rule instead of classic UCT (detailed in the last section), which dynamically adjusts exploration–exploitation to suit the specific characteristics of LLM output distributions.

\paragraph{Expansion.}
This phase directly utilizes the strong model’s pre-trained parameters to generate new child nodes of the selected leaf node. 
To reflect the strong model's inherent tendencies (corresponding to the first term in $\mathcal{G}$), we first identify the Top-$N$ most probable tokens based on its $\pi_{\text{strong}}$ predicted distribution at the current state $s$.
We then generalize the token-level generation to a more flexible step-level generation, enhancing search efficiency.
For diverse exploration, $K$ distinct tokens are randomly selected from these Top-$N$ candidates to initiate the formation of new chunks.
Each selected token then forms a new chunk $y_{1:L}$ of length $L$, which is subsequently concatenated to the current sequence $\mathbf{y}'$ to yield the new state $s'$. 
This design allows for flexible setting of the step size $L$ to adapt different tasks:
\begin{itemize}
    \item Fine-grained decision-making: When $L=1$, MCTS performs precise token-by-token decisions, achieving high-accuracy control over alignment at each step.
    \item High-level branching: When $L>1$, MCTS expands a short sequence at a time, effectively reducing the tree depth, thereby enabling more efficient and higher-level exploration in high-dimensional spaces.
\end{itemize}

\paragraph{Evaluation.} Each newly generated node $s'$ (representing the partial sequence $(\mathbf{x}, \mathbf{y}' \circ y_{1:L})$) undergoes a crucial evaluation by computing the proxy value, serving as its immediate feedback. 
We promote the proxy value function of Eq.~\ref{eq:weak_reward_model}:
\begin{equation}
V_{\text{proxy}}(\mathbf{x}, \mathbf{y}' \circ y_{1:L}) = \log\frac{\pi_{\text{weak}}^*(\mathbf{y}' \circ y_{1:L}|\mathbf{x})}{\pi_{\text{weak}}^{\text{ref}}(\mathbf{y}' \circ y_{1:L}|\mathbf{x})}.
\end{equation}
The effective value of the node is initialized with this $V_{\text{proxy}}$, serving as the immediate reward $R(s')$ for that node in the tree. 
If a child node meets the stopping condition (e.g., reaching maximum length or generating \verb|EOS| token), its return $R(s')$ is explicitly set to $-\infty$ to prevent its re-selection in subsequent MCTS iterations.
These termination nodes will remain candidates for evaluation in Stage 2.

\paragraph{Backpropagation.} 
The $R(s')$ value obtained from the newly simulated node is precisely back-propagated along its path to the root node, with the each ancestor state's visit count incrementally increasing. 
Crucially, the the parent state's return $R(s_{\text{p}})$ updates to the maximum of all observed child node returns, following the formula:
\begin{equation}
R(s_{\text{p}}) \leftarrow \max(R(s_{\text{c}})).
\end{equation}
This fundamental design makes MCTS propagate maximum rather than typical average returns, treating LLM alignment as an optimal search for the single highest-reward sequence rather than a long-term average returns like in an adversarial game~\cite{silver2016mastering}. 
It ensures that MCTS preserves high-value trajectories, effectively prunes sub-optimal branches, and concentrates computational resources on the most promising paths.

\begin{figure*}[!ht]
  \centering
  \includegraphics[width=0.9\linewidth]{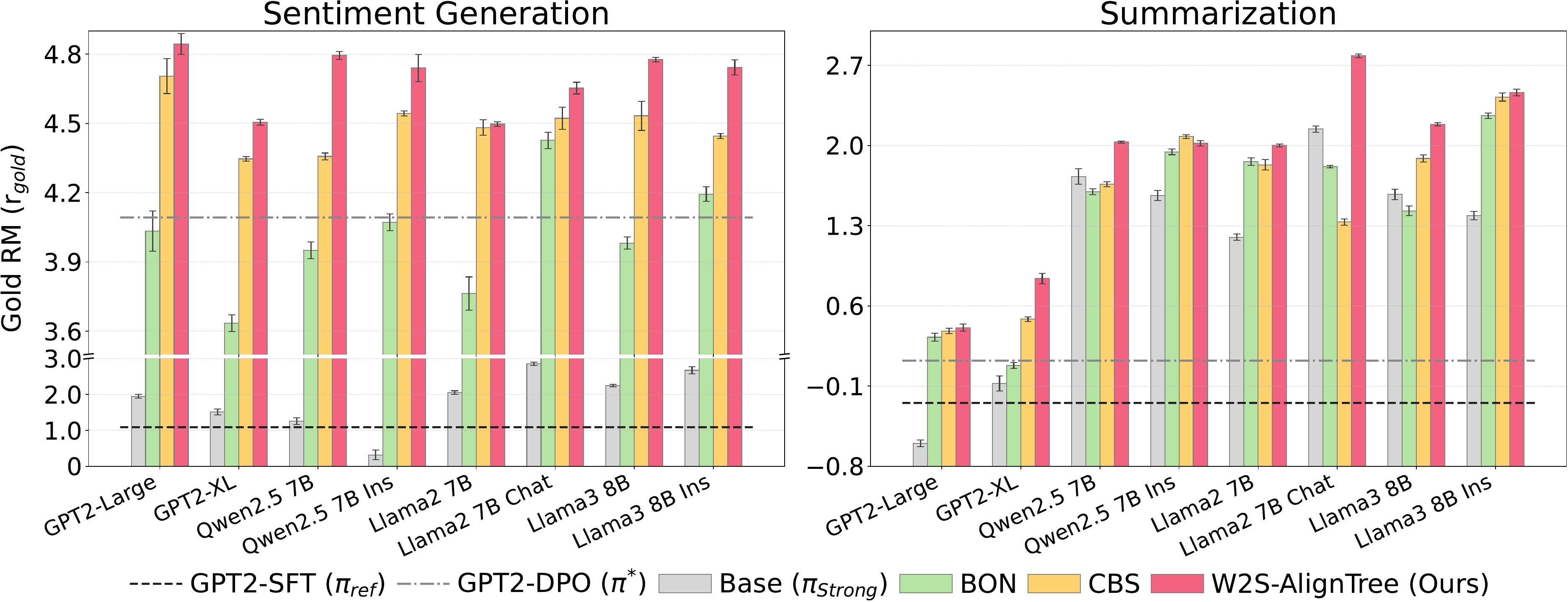}
  \caption{Alignment performance across sentiment generation and summarization. W2S-AlignTree consistently outperforms strong baselines by enabling inference-time alignment with weak model guidance, achieving higher $r_{\mathrm{gold}}$ across diverse LLMs.
  We report mean rewards ($\pm$ standard deviations) across three random seeds.
  Appendix C.1 provides detailed numerical results.}
  \label{fig:combined_results}
\end{figure*}

\subsection{Stage 2: Best Candidate Decision}

After MCTS iterations, we obtain a generative search tree built with intermediate step-level $V_{\text{proxy}}$, but a final decision still requires global evaluation.
Stage 2 aims to identify the best-aligned complete response among all explored nodes by a robust strategy.
As shown in Fig.~\ref{fig:main} (b), a recursive Depth-First Search (\textit{DFS}) first collects all complete sequences from root to terminal nodes.
Among them, we identify candidates whose penultimate nodes have the highest MCTS-evaluated returns, reflecting paths most promising for alignment.
For every candidate, its final global alignment We compute the final global alignment score for each candidate according to:
\begin{equation}
\label{eq:full_reward_stage2}
    r_{\text{proxy}}(\mathbf{x},\mathbf{y}) = \frac{\log\pi_{\text{weak}}^*(\mathbf{y}|\mathbf{x})}{\log\pi_{\text{weak}}^{\text{ref}}(\mathbf{y}|\mathbf{x})}.
\end{equation}
This score, which matches the full sentence-level alignment objective in Eq.~\ref{eq:value or reward with y}, is then used to re-rank the candidates.
The candidate with the highest global alignment score will be ultimately decided as the final aligned output.
If no terminal node is found during MCTS iterations (e.g., due to insufficient search budget or an extremely difficult problem), we design a fallback mechanism that selects the node with the highest MCTS return, guaranteeing stability and consistent output.
Stage 2 integrates both step-level guidance and sequence-level evaluation, leading to a substantial improvement in the reliability, fidelity and quality of the results.

\subsection{Entropy-Aware PUCT Selection Rule}
\label{sec:ea-puct}
During the MCTS selection phase, we design an EA-PUCT rule to adaptively balance exploration and exploitation.
The standard UCT guides exploration by balancing node visit counts and the average value of all historical returns, and PUCT combines prior probabilities from a policy network into UCT.
However, the output distribution of $\pi_{\text{strong}}$ often exhibits ``peak" effects, causing MCTS to converge prematurely to locally optimal solutions at the expense of diversity and the discovery of better alternatives.
To overcome this, we draw on the concept of information entropy into the exploration bonus and propose EA-PUCT, which is defined as:
\begin{equation}
\label{eq:ea-puct}
{E\text{-}PU}(s) = R(s) + c \cdot P(s) \cdot \frac{\sqrt{N(s_{\text{p}})}}{1+N(s)} \cdot (1 + w \cdot H(s)),
\end{equation}
where $P(s)$ is the prior probability of the strong model's action leading to state $s$ from the parent $s_{\text{p}}$, $H(s)$ denotes the information entropy, $c$ and $w$ are coefficients.
Note that $R(s)$ here denotes the immediate return but not an average, since our task is framed as an optimal search.
For a generated chunk of length $L$, $P(s)$ is precisely defined as the geometric mean of the token-level probabilities: $P(s) = \exp\left(\frac{1}{L}\sum_{t=1}^{L} \log p(y_t|s_{\text{p}},y_{<t})\right)$. This better penalizes low-probability tokens, boosting robust path exploration.

The information entropy $H(s)$ measures the uncertainty of the strong model's output distribution and is defined as:
\begin{equation}
H(s) = -{\sum}_{a}P(s,a) \cdot \log P(s,a).
\end{equation}
By incorporating $(1+w \cdot H(s))$ as an uncertainty-aware guide for the exploration bonus, EA-PUCT endows the search with information-gain consciousness:
\begin{itemize}
    \item When the entropy is large, it signals that the model’s next-token distribution is highly uncertain; the term $(1 + w \cdot H(s))$ then inflates the exploration bonus, urging MCTS to delve more deeply into diverse trajectories.  
    \item Conversely, a low entropy indicates that the model is confident about the best action; the exploration bonus is suppressed, and the search shifts its emphasis toward exploiting the already-identified high-reward paths.  
\end{itemize}

This mechanism effectively mitigates the premature convergence in MCTS for LLM alignment, markedly enhancing the ability to explore diverse candidate answers in a complex generation space while preventing the excessive randomness that would arise from naive entropy maximization.

\begin{table*}[!htbp]
\centering
\ifdefined\colw
\else
    \newlength\colw
\fi

\setlength\colw{0.10\textwidth}
\setlength\tabcolsep{2pt}
\renewcommand\arraystretch{0.85}

\begin{tabular}{l*{8}{>{\centering\arraybackslash}p{\colw}}}
\toprule
& {\normalsize Qwen2.5-7B} & {\normalsize Qwen2.5-7B-Instruct} & {\normalsize Llama3-8B} & {\normalsize Llama3-8B-Instruct} &
  {\normalsize Llama2-7b-hf} & {\normalsize Llama2-7b-chat-hf} & {\normalsize tulu2-7b} & {\normalsize tulu2-7b-dpo} \\ \midrule

\multicolumn{9}{c}{{\normalsize \makecell{{\normalsize \textbf{Gold Reward Model: oasst-rm-2-pythia-6.9b}} \\    {\normalsize Llama3.2-1B-Instruct / Llama3.2-1B: 0.64 / -0.82}}}}
\\ \midrule
{\normalsize Base}  & {\normalsize 0.90} & {\normalsize 1.45} & {\normalsize -0.68} & {\normalsize 0.81} & {\normalsize -0.75} & {\normalsize 0.79} & {\normalsize -0.13} & {\normalsize 0.52} \\
{\normalsize BoN}  & {\normalsize 0.91} & {\normalsize 1.49} & {\normalsize -1.28} & {\normalsize 0.90} & {\normalsize -1.22} & {\normalsize 0.95} & {\normalsize 0.46}  & {\normalsize 0.57} \\
{\normalsize CBS}   & {\normalsize 0.75} & {\normalsize \textbf{1.67}} & {\normalsize -0.56} & {\normalsize \textbf{1.13}} & {\normalsize -0.52} & {\normalsize 0.44} & {\normalsize -0.73} & {\normalsize 0.64} \\
{\normalsize W2S-AT (Ours)}  & {\normalsize \textbf{1.33}} & {\normalsize 1.51} & {\normalsize \textbf{-0.10}} & {\normalsize 0.97} & {\normalsize \textbf{-0.50}} & {\normalsize \textbf{1.01}} & {\normalsize \textbf{0.72}}  & {\normalsize \textbf{0.75}} \\ \midrule

\multicolumn{9}{c}{{\normalsize \makecell{{\normalsize \textbf{Gold Reward Model: UltraRM-13b}} \\  {\normalsize  Llama3.2-1B-Instruct / Llama3.2-1B: -6.13 / -10.01}}}}
\\ \midrule
{\normalsize Base} & {\normalsize -3.05} & {\normalsize 0.62} & {\normalsize -9.15} & {\normalsize -3.45} & {\normalsize -9.56} & {\normalsize -3.89} & {\normalsize -6.48} & {\normalsize -4.80} \\
{\normalsize BoN}   & {\normalsize -1.13} & {\normalsize 0.57} & {\normalsize -10.40} & {\normalsize -2.19} & {\normalsize -10.52} & {\normalsize -3.52} & {\normalsize -5.59} & {\normalsize -4.24} \\
{\normalsize CBS}    & {\normalsize -2.21} & {\normalsize \textbf{1.43}} & {\normalsize -9.53} & {\normalsize \textbf{-1.44}} & {\normalsize -8.72} & {\normalsize -3.44} & {\normalsize -8.04} & {\normalsize -4.12} \\
{\normalsize W2S-AT (Ours)}  & {\normalsize \textbf{-1.02}} & {\normalsize 1.01} & {\normalsize \textbf{-7.56}} & {\normalsize -1.96} & {\normalsize \textbf{-8.49}} & {\normalsize \textbf{-3.29}} & {\normalsize \textbf{-4.37}} & {\normalsize \textbf{-3.39}} \\
\bottomrule
\end{tabular}
\caption{
Evaluating instruction-following on \texttt{OASST1} using W2S-AlignTree (W2S-AT) and representative baselines.
We use \texttt{Llama3.2-1B-Instruct} and \texttt{Llama3.2-1B} as weak guidance models,
and both \texttt{oasst-rm-2-pythia-6.9b} and \texttt{UltraRM-13b} as gold reward models.
\textbf{Bold} indicates the best $r_{\mathrm{gold}}$ score in each column.
}
\label{tab:llama3_1b_ins}
\end{table*}

\section{Experiments}
\subsection{Experimental Setup}
We evaluate W2S-AlignTree’s ability to use a weak LLM to guide larger models across diverse tasks, model families and scales.
More details are provided in Appendix B.

\paragraph{Task Designs.}
We evaluate three progressively challenging language tasks using standard datasets: controlled-sentiment generation on \texttt{IMDB}~\cite{maas2011learning}, summarization on \texttt{TL;DR}~\cite{stiennon2020learning}, and instruction-following on \texttt{OASST1}~\cite{kopf2023openassistant}.
For sentiment generation and summarization, we use a supervised fine‑tuned \texttt{GPT2} model $\pi_{\text{weak}}^*$ (124M parameters) and its DPO‑tuned variant $\pi_{\text{weak}}^{\text{ref}}$ to emulate the target behaviors, and jointly use these models to guide a series of larger LLMs.
For the more demanding instruction-following, we employ off‑the‑shelf models and their untuned counterparts (e.g., \texttt{Llama-3.2-1B-Instruct} and \texttt{Llama-3.2-1B}) as weak guidance signals to steer several powerful LLMs, thereby illustrating that weak models remain universally applicable without any task‑specific tuning.

\paragraph{Baselines and Evaluation.}
We evaluate W2S-AlignTree against several baselines with the same score of Eq.~\ref{eq:full_reward_stage2} for fair comparison:
(1) Base model ($\pi_{\text{strong}}$): we employ regular generation of the frozen LLMs.  
(2) Best‑of‑N (BoN)~\cite{touvron2023llama}: a post‑hoc selection method that chooses the candidate with the highest score from $N$ generated outputs.  
(3) Chunk-level Beam Search (CBS)~\cite{zhou2024weak}: an inference‑time alignment technique that dynamically integrates reward signals into the beam search process.  
Whenever applicable, we contrast inference‑time methods with DPO‑tuning of the large LLMs, detailed in Appendix C.1.
Following common practices~\cite{rafailov2023direct,zhu2024weak}, we adopt the gold reward‑model score $r_{\mathrm{gold}}$, computed by a high‑fidelity pre-trained reward model, to assess alignment quality (higher indicates better).

\subsection{Experimental Results}
Experimental results show that W2S‑AlignTree consistently and significantly surpasses strong baselines, achieving superior fine‑grained alignment across tasks.

\paragraph{Sentiment Generation \& Summarization.}
Results under W2S-AlignTree consistently outperforms strong baselines in both controlled-sentiment generation and summarization, guided by paired \texttt{GPT2} models. 
We mainly evaluate across eight models: in-family \texttt{GPT2-Large} (774M) and \texttt{GPT2-XL} (1.5B), as well as cross-family mainstream base models such as \texttt{Llama2-7b-hf}, \texttt{Llama3-8B} and \texttt{Qwen2.5-7B}, alongside their aligned versions.

For the controlled-sentiment generation~(Fig.~\ref{fig:combined_results}, Left), W2S-AlignTree consistently elevates the $r_{\mathrm{gold}}$ across all models. 
In particular, it achieves $r_{\mathrm{gold}}$ of $4.79$ on \texttt{Qwen2.5-7B}, representing a significant $10.04\%$ improvement over the second-best CBS with $4.36$.
It also yields over $5\%$ gains on models such as \texttt{Llama3-8B} and \texttt{Llama3-8B-Instruct}, highlighting its effectiveness in aligning generated content with target sentiment.
In summarization~(Fig.~\ref{fig:combined_results}, Right), all models employing W2S-AlignTree again show robust and stable performance gains except \texttt{Qwen2.5-7B-Instruct}, which is also comparable to baselines.
For instance, \texttt{GPT2-Large} and \texttt{GPT2-XL} initially exhibit negative $r_{\mathrm{gold}}$ scores of $-0.60$ and $-0.08$ due to their weaker inherent capabilities.
Under W2S-AlignTree, these scores rise to $0.41$ and $0.84$, with \texttt{GPT2-XL}’s improvement clearly outperforming the baselines and demonstrating enhanced factual and semantic consistency.
Moreover, W2S-AlignTree achieved a high $r_{\mathrm{gold}}$ of $2.78$ on the \texttt{Llama-2-7b-chat-hf}, marking a substantial $29.84\%$ improvement compared to the second best direct inference model of $2.14$.
Generally, CBS greedily aggregates alignment signals at fixed beam width per chunk; while W2S-AlignTree treats signals as explicitly backed-up global values and adaptively explores high-reward long-sequence branches via EA-PUCT, mitigating local-optimum and credit-assignment issues. 
So W2S-AlignTree enhances alignment across both aligned and unaligned models by leveraging weak model's guidance, ultimately yielding semantically accurate and user-intended outputs.

\paragraph{Instruction-Following.}
W2S-AlignTree also shows clear advantages in instruction-following across diverse models without task-specific training, highlighting its generality and practical utility.
As shown in Tab.~\ref{tab:llama3_1b_ins}, we evaluate W2S-AlignTree using \texttt{Llama3.2-1B-Instruct} and its untuned \texttt{Llama3.2-1B} as weak guidance.
To assess more comprehensively, we employ two distinct reward models as gold evaluators: \texttt{oasst-rm-2-pythia-6.9b}~\cite{kopf2023openassistant}, specifically fine-tuned on \texttt{OASST1} to reflect task-specific alignment, and \texttt{UltraRM-13b}~\cite{cui2023ultrafeedback}, a general-purpose reward model for instruction evaluation across domains.
Results indicate that W2S-AlignTree consistently achieves the highest, or occasionally second-highest $r_{\mathrm{gold}}$ across most model configurations, demonstrating its stability and strong cross-model generalizability. 
For instance, when applied to \texttt{Qwen2.5-7B}, it raises the score to $1.33$, significantly outperforming the next-best method (BoN, $0.91$). 
Under the more stringent UltraRM evaluation, it improves \texttt{Llama3-8B}'s score from $-9.53$ to $-7.56$.
Additional results in Appendix C.2 confirm that W2S-AlignTree remains effective when guided by  other smaller models, including but not limited to \texttt{Qwen2.5-0.5B}, demonstrating broad adaptability.
Overall, W2S-AlignTree provides a robust and scalable approach for inference-time alignment, enabling strong LLMs to generate high-quality, instruction-aligned outputs at low cost.

\begin{table}[!htbp]
\centering
\small
\renewcommand\arraystretch{0.9}
\begin{tabular}{lcccc}
\toprule
 & \multicolumn{2}{c}{{\normalsize Sentiment Gen.}} & \multicolumn{2}{c}{{\normalsize Summarization}} \\
\cmidrule(lr){2-3}\cmidrule(lr){4-5}
 & GPT-XL & Llama3-8B & GPT-XL & Llama3-8B \\
\midrule
N-UCT   & {\normalsize 3.67} & {\normalsize 3.30} & {\normalsize 0.67} & {\normalsize 1.40} \\
RT-UCT  & {\normalsize 4.09} & {\normalsize 3.89} & {\normalsize 0.67} & {\normalsize 1.63} \\
RT-PUCT & {\normalsize 4.39} & {\normalsize 4.57} & {\normalsize 0.64} & {\normalsize 2.12} \\
CMB     & {\normalsize 3.47} & {\normalsize 3.29} & {\normalsize 0.52} & {\normalsize 1.46} \\
MMB     & {\normalsize 4.16} & {\normalsize 4.66} & {\normalsize 0.61} & {\normalsize 1.85} \\
\textbf{W2S-AT}
& \textbf{\normalsize 4.51} & \textbf{\normalsize 4.80}
& \textbf{\normalsize 0.84} & \textbf{\normalsize 2.18} \\

\bottomrule
\end{tabular}
\caption{Ablation of key components in W2S-AlignTree cross tasks with \texttt{Llama3-8B} and \texttt{GPT-XL}.}
\label{tab:ablation_results}
\end{table}

\subsection{Ablation Studies}

We conduct a series of ablation studies to assess the contributions of W2S-AlignTree’s key components and hyperparameter settings to alignment performance and robustness.

\paragraph{Key Technical Components.}
To clarify the impact of each main component of W2S-AlignTree on its alignment, we design five variants for ablation experiments: (1) Naive UCT (N-UCT): basic UCT with average historical return backpropagation. (2) Real-Time UCT (RT-UCT): backpropagating the current return to avoid averaging latency. (3) Real-Time PUCT (RT-PUCT): incorporating strong model priors in selection and real-time return in backpropagation. (4) Child Mean Backpropagation (CMB): backpropagating the mean of all child node returns. (5) Mixed Mean Backpropagation (MMB): backpropagating the average mix of maximum and mean child returns.
The comparative experimental results on in-family \texttt{GPT2-XL} and cross-family \texttt{Llama3-8B} are presented in Tab.~\ref{tab:ablation_results}. 
The complete W2S-AlignTree, with maximum immediate return backpropagation and EA-PUCT strategy, consistently performs best. Ablation studies reveal that removing key components significantly degrades performance: specifically, the maximum return backpropagation (CMB and MMB) is crucial for preserving optimal paths, while the absence of prior or information entropy (N-UCT, RT-UCT) hinders strategic guidance and adaptive adjustment. RT-PUCT further demonstrates the importance of entropy intervention in preserving the model's exploration capability. These findings validate the effectiveness of W2S-AlignTree's core designs, particularly the key role of maximum return propagation and entropy-enhanced prior guidance in achieving fine-grained alignment.

\begin{figure}[!ht]
  \centering
  \includegraphics[width=0.88\linewidth]{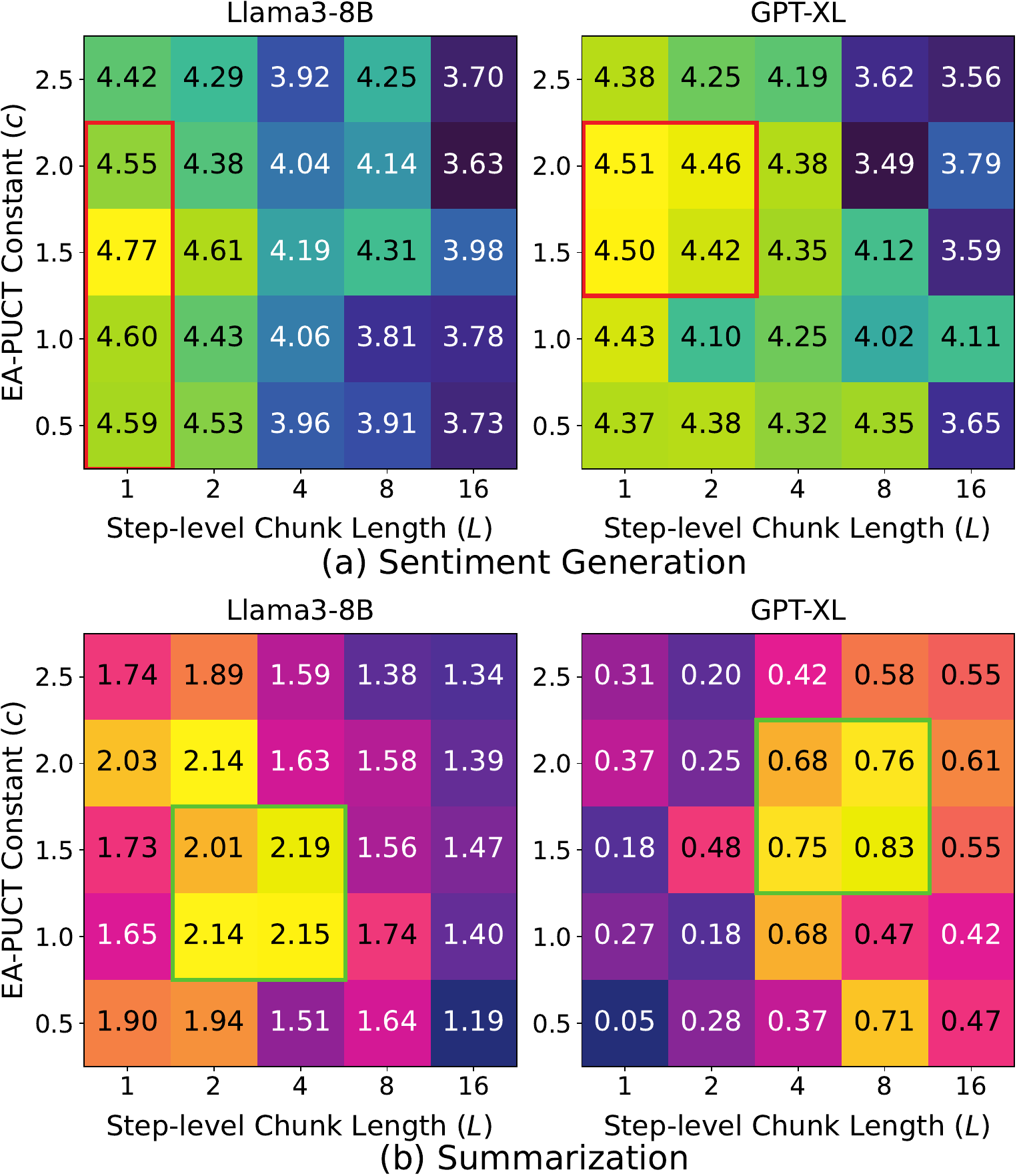}
  \caption{Hyperparameter analysis of W2S-AlignTree to chunk length $L$ and exploration coefficient $c$ across tasks. The areas with better performance are boxed.}
  \label{fig:para_results}
\end{figure}

\paragraph{Hyperparameter Sensitivity.}
We further analyze the sensitivity of W2S-AlignTree's hyperparameters, mainly focusing on step-level chunk length $L$ and the EA-PUCT exploration constant $c$ here, as visualized in Fig.~\ref{fig:para_results}. 
For $L$, we observe task-specific preferences. 
For controlled-sentiment generation, which demands fine-grained controllability, smaller $L$ (e.g., $L=1$ as token-level decisions) generally yields optimal performance. 
While for summarization, slightly longer $L$ prove more suitable, emphasizing semantic coherence through wider context capture. 
Regarding $c$, we find that model performance is most stable and optimal when $c \in [1.0, 2.0]$, balancing the trade-off between exploration of uncertain paths and exploitation of known high-return branches.
Extremely small values of $c$ lead to myopic behavior and early convergence, while larger values introduce unnecessary variance and degrade reliability. 
And quantitative results in Fig.~\ref{fig:para_results} collectively indicate that W2S-AlignTree consistently achieves high performance under more than one configuration, highlighting its strong stability and robustness. 
Additional analyses for other parameters are detailed in Appendix C.3.

\section{Conclusion}
We propose \textbf{W2S-AlignTree}, a pioneering plug-and-play inference-time alignment framework that overcomes the high costs and limited controllability of training-time alignment. 
W2S-AlignTree is the first to systematically integrate MCTS with W2S paradigm, recasting LLM alignment as an optimal search problem balancing exploration and exploitation in a generative tree. 
By leveraging dynamic, step-level signals from a weak model as alignment proxies and introducing the Entropy-Aware PUCT selection rule, W2S-AlignTree achieves fine-grained guidance over the strong model's generation without modifying its parameters. 
Comprehensive experiments show that W2S-AlignTree consistently enhances alignment quality on tasks such as controlled-sentiment generation, summarization and instruction-following. 
Superior results across diverse model families, scales, and hyperparameters further highlight the universality and robustness.
In conclusion, W2S-AlignTree reveals the effectiveness of integrating MCTS with the W2S paradigm. By enabling more fine-grained and dynamic control over LLM behavior, it offers a scalable alignment solution, opening up a new perspective for practically balancing the safety, controllability and utility of LLMs.

\section{Acknowledgments}
This research is supported in part by the National Key Research and Development Program of China (Grant No. 2023ZD0121300), and in part by the National Natural Science Foundation of China (Grant No. 62495092, 62088102).

\bibliography{aaai2026}

\clearpage
\appendix

\section{Method Details}
\subsection{Reproducibility}
To facilitate review and reproducibility, experimental setup, code and data supporting W2S-AlignTree are publicly accessible at \textcolor{purple}{https://github.com/alexzdy/W2S-AlignTree}.

\subsection{Token-level Reward Derivation}
To obtain a dense reward, we decompose the sequence-level alignment reward $r(\mathbf{x}, \mathbf{y})$. Using the closed-form solution of $\pi^*(\mathbf{y}|\mathbf{x})$ and the chain rule, it can be expressed as:
\begin{equation}
\label{eq:reward}
r(\mathbf{x}, \mathbf{y}) = \beta \log \frac{\pi^*(\mathbf{y}|\mathbf{x})}{\pi_{\text{ref}}(\mathbf{y}|\mathbf{x})}.
\end{equation}

By applying the chain rule of probability to both the numerator and the denominator, we can express the ratio of sequence probabilities as a product of token-level probability ratios:
\begin{equation}
\frac{\pi^*(\mathbf{y}|\mathbf{x})}{\pi_{\text{ref}}(\mathbf{y}|\mathbf{x})} = \frac{\prod_{t=1}^{|\mathbf{y}|} \pi^*(y_t | \mathbf{x}, \mathbf{y}_{<t})}{\prod_{t=1}^{|\mathbf{y}|} \pi_{\text{ref}}(y_t | \mathbf{x}, \mathbf{y}_{<t})} = \prod_{t=1}^{|\mathbf{y}|} \frac{\pi^*(y_t | \mathbf{x}, \mathbf{y}_{<t})}{\pi_{\text{ref}}(y_t | \mathbf{x}, \mathbf{y}_{<t})}.
\end{equation}

Substituting this into Eq.~\ref{eq:reward} and using the property of logarithms ($\log(AB) = \log A + \log B$), we can convert the product into a sum:
\begin{equation}
\begin{aligned}
r(\mathbf{x}, \mathbf{y}) &= \beta \log \left( \prod_{t=1}^{|\mathbf{y}|} \frac{\pi^*(y_t | \mathbf{x}, \mathbf{y}_{<t})}{\pi_{\text{ref}}(y_t | \mathbf{x}, \mathbf{y}_{<t})} \right)   \\
&= \beta \sum_{t=1}^{|\mathbf{y}|} \log \left( \frac{\pi^*(y_t | \mathbf{x}, \mathbf{y}_{<t})}{\pi_{\text{ref}}(y_t | \mathbf{x}, \mathbf{y}_{<t})} \right),
\end{aligned}
\end{equation}
where $\mathbf{y}_{<t}$ represents the prefix up to token $t-1$, $y_{|\mathbf{y}|}$ is the \verb|EOS| token, and $Z(\mathbf{x})$ is omitted as it does not influence the objective. This converts sparse sequence-level rewards into a continuous stream of token-level evaluations.

\subsection{Step-level Value Function Derivation}
We model the generation process as a token-level MDP $\mathcal{M} = (\mathcal{S}, \mathcal{A}, f, r)$.

\begin{itemize}
    \item The state $s_t = (\mathbf{x}, \mathbf{y}_{<t})$ is a partial prefix $\mathbf{y}'$, with the initial state being $s_0 = (\mathbf{x}, \varnothing)$.
    \item The action $a_t = y_t$ is the generated token.
    \item The reward function $r(s_t, a_t)$ is sparse, meaning a non-zero reward is received only upon the completion of a full sequence. Specifically, the reward equals the sequence reward $r(\mathbf{x}, \mathbf{y})$ only at the final step when the action $a_t = \verb|EOS|$ is the end-of-sequence token, and is 0 for all preceding steps.
\end{itemize}

The optimal policy $\pi^*$ and the optimal soft value functions $V^*$ and $Q^*$ satisfy the following soft Bellman equations~\cite{rafailov2024r,zhou2024weak}:
\begin{equation}
\label{eq:soft-bellman-q}
Q^*(s_t, a_t) = r(s_t, a_t) + V^*(s_{t+1}),
\end{equation}
\begin{equation}
\label{eq:soft-bellman-v}
V^*(s_t) = \beta \log \sum_{a'} \pi_{\text{ref}}(a'|s_t) \exp\left( \frac{1}{\beta} Q^*(s_t,a') \right).
\end{equation}
The optimal policy has a closed-form solution:
\begin{equation}
\label{eq:log-ratio}
\log \frac{\pi^*(a_t | s_t)}{\pi_{\text{ref}}(a_t | s_t)} = \frac{1}{\beta} \left( Q^*(s_t, a_t) - V^*(s_t) \right).
\end{equation}
From Eq.~\ref{eq:soft-bellman-v}, we define the partition function $Z(s_t)$ as $Z(s_t) = \sum_{a'} \pi_{\text{ref}}(a'|s_t) \exp\left( \frac{1}{\beta} Q^*(s_t,a') \right)$, which implies a direct relationship between the value function and the log of the partition function:
\begin{equation}
\label{eq:v-z-relation}
V^*(s_t) = \beta \log Z(s_t).
\end{equation}
This relationship is universal for any state $s_t$, including the initial state $s_0$, thus $V^*(\mathbf{x}) = \beta \log Z(\mathbf{x})$.

Substituting Eq.~\ref{eq:soft-bellman-q} into Eq.~\ref{eq:log-ratio}, we obtain the single-step decomposition:
\begin{equation}
\label{eq:single-step-decomposition}
\beta \log \frac{\pi^*(a_t | s_t)}{\pi_{\text{ref}}(a_t | s_t)} = r(s_t, a_t) + V^*(s_{t+1}) - V^*(s_t).
\end{equation}
We sum this expression from $t=0$ to $H-1$, where $H$ is the length of the generated sequence or prefix $\mathbf{y}'$. The left-hand side is the log-probability ratio for the entire sequence or prefix, while the right-hand side is a telescoping sum plus the total reward:
\begin{equation}
\label{eq:telescoping-sum}
\beta \log \frac{\pi^*(\mathbf{y}'|\mathbf{x})}{\pi_{\text{ref}}(\mathbf{y}'|\mathbf{x})} = \sum_{t=0}^{H-1} r(s_t, a_t) + V^*(s_H) - V^*(s_0).
\end{equation}
Here, $V^*(s_0)$ is the value of the initial state $s_0 = (\mathbf{x}, \varnothing)$, which we denote as $V^*(\mathbf{x})$.

Then, we analyze two cases based on whether the sequence is complete:
\begin{itemize}
    \item \textbf{Incomplete prefix ($\mathbf{y}' \neq \mathbf{y}$)}:
    In this case, no \verb|EOS| token has been generated. The total reward is zero, and the final state is $s_H = (\mathbf{x}, \mathbf{y}')$, with its value being $V^*(\mathbf{x}, \mathbf{y}')$.
    \begin{equation}
    \label{eq:incomplete-prefix}
    \beta \log \frac{\pi^*(\mathbf{y}'|\mathbf{x})}{\pi_{\text{ref}}(\mathbf{y}'|\mathbf{x})} = V^*(\mathbf{x}, \mathbf{y}') - V^*(\mathbf{x}).
    \end{equation}
    \item \textbf{Complete sequence ($\mathbf{y}' = \mathbf{y}$)}:
    In this case, the last action was the \verb|EOS| token. The total reward is the sequence reward $r(\mathbf{x}, \mathbf{y})$, and the final state $s_H$ is a terminal state with a value of zero.
    \begin{equation}
    \label{eq:complete-sequence}
    \beta \log \frac{\pi^*(\mathbf{y}|\mathbf{x})}{\pi_{\text{ref}}(\mathbf{y}|\mathbf{x})} = r(\mathbf{x}, \mathbf{y}) - V^*(\mathbf{x}).
    \end{equation}
\end{itemize}
In summary, we have derived a general decomposition formula that includes a baseline value from the initial state. As shown in Eq.~\ref{eq:v-z-relation}, this baseline is $V^*(\mathbf{x}) = \beta \log Z(\mathbf{x})$, where $Z(\mathbf{x})$ is the partition function for the initial state. In practice, when comparing the log-probability ratios of different sequences or prefixes, we are interested in their relative values. Since the term $-V^*(\mathbf{x})$ depends only on the initial state $\mathbf{x}$ and is constant for all generated sequences from that prompt, it can be treated as a common baseline, which is consistent with the conclusion regarding $Z(\mathbf{x})$. 
By an abuse of notation, we can effectively define this constant as zero to simplify the expression. 
If the complete output is $\mathbf{y}$, this leads to the desired concise form:
\begin{equation}  
\label{eq:value-final-form}
    \beta\log\frac{\pi^*(\mathbf{y}'|\mathbf{x})}{\pi_{\text{ref}}(\mathbf{y}'|\mathbf{x})} = 
\begin{cases} V^*(\mathbf{x},\mathbf{y}'), & \text{if } \mathbf{y}'\neq\mathbf{y} \\ 
r(\mathbf{x},\mathbf{y}), & \text{if } \mathbf{y}'=\mathbf{y}. \end{cases}
\end{equation}

\subsection{Interpretation of Weak-to-Strong Proxy Mapping}
In the W2S-AlignTree framework, we utilize the alignment signal of a weak model as a proxy to guide the selection of output sequences from a strong model. We will theoretically prove that, under appropriate assumptions, the weak model's proxy reward is consistent with (or at least monotonically related to) the strong model's true alignment reward, thereby demonstrating the validity and effectiveness of using the weak model as a proxy.

First, let us define our terms and assumptions. The strong model's optimal policy is $\pi_{\text{strong}}^*(\mathbf{y}|\mathbf{x})$, which is designed to optimize the true sequence-level alignment reward:
\begin{equation}
\label{eq:strong-reward}
r(\mathbf{x}, \mathbf{y}) = \beta\log\frac{\pi_{\text{strong}}^*(\mathbf{y}|\mathbf{x})}{\pi_{\text{strong}}^{\text{ref}}(\mathbf{y}|\mathbf{x})}.
\end{equation}
We define the weak model's policies as the unaligned weak model $\pi_{\text{weak}}^{\text{ref}}(\mathbf{y}|\mathbf{x})$ and the aligned weak model $\pi_{\text{weak}}^*(\mathbf{y}|\mathbf{x})$. The proxy reward is defined as:
\begin{equation}
\label{eq:weak-reward}
R_{\text{weak}}(\mathbf{x}, \mathbf{y}) = \beta\log\frac{\pi_{\text{weak}}^*(\mathbf{y}|\mathbf{x})}{\pi_{\text{weak}}^{\text{ref}}(\mathbf{y}|\mathbf{x})}.
\end{equation}

To establish a connection, we propose two assumptions:
\begin{itemize}
    \item \textbf{Assumption 1:} The weak model can capture the primary alignment features that the strong model focuses on, such that its aligned distribution is proportional to the strong model's optimal distribution raised to the power of $\alpha$:
    \begin{equation}
    \label{eq:assumption1_new}
    \pi_{\text{weak}}^*(\mathbf{y}|\mathbf{x}) \propto [\pi_{\text{strong}}^*(\mathbf{y}|\mathbf{x})]^{\alpha},
    \end{equation}
    where $\alpha > 0$ is a temperature or scaling factor. This implies that there exists a normalization constant $C_A > 0$ such that $\pi_{\text{weak}}^*(\mathbf{y}|\mathbf{x}) = C_A \cdot [\pi_{\text{strong}}^*(\mathbf{y}|\mathbf{x})]^{\alpha}$.

    \item \textbf{Assumption 2:} The unaligned version of the weak model is structurally similar to the strong model's reference model, such that its reference distribution is proportional to the strong model's reference distribution raised to the power of $\alpha$:
    \begin{equation}
    \label{eq:assumption2_new}
    \pi_{\text{weak}}^{\text{ref}}(\mathbf{y}|\mathbf{x}) \propto [\pi_{\text{strong}}^{\text{ref}}(\mathbf{y}|\mathbf{x})]^{\alpha},
    \end{equation}
    where $\alpha$ is the same as in Assumption 1. This implies that there exists a normalization constant $C_B > 0$ such that $\pi_{\text{weak}}^{\text{ref}}(\mathbf{y}|\mathbf{x}) = C_B \cdot [\pi_{\text{strong}}^{\text{ref}}(\mathbf{y}|\mathbf{x})]^{\alpha}$.
\end{itemize}

Based on these two new assumptions, we can derive the relationship between the proxy and true rewards. We substitute Assumption 1 and Assumption 2 into the definition of the proxy reward Eq.~\ref{eq:weak-reward}:
\begin{equation}
\begin{aligned}
R_{\text{weak}}(\mathbf{x}, \mathbf{y}) &= \beta\log\frac{\pi_{\text{weak}}^*(\mathbf{y}|\mathbf{x})}{\pi_{\text{weak}}^{\text{ref}}(\mathbf{y}|\mathbf{x})} \\
&= \beta\log\frac{C_A \cdot [\pi_{\text{strong}}^*(\mathbf{y}|\mathbf{x})]^{\alpha}}{C_B \cdot [\pi_{\text{strong}}^{\text{ref}}(\mathbf{y}|\mathbf{x})]^{\alpha}} \\
&= \beta\log\left( \frac{C_A}{C_B} \cdot \left[ \frac{\pi_{\text{strong}}^*(\mathbf{y}|\mathbf{x})}{\pi_{\text{strong}}^{\text{ref}}(\mathbf{y}|\mathbf{x})} \right]^{\alpha} \right) \\
&= \beta \left( \log\frac{C_A}{C_B} + \log\left( \left[ \frac{\pi_{\text{strong}}^*(\mathbf{y}|\mathbf{x})}{\pi_{\text{strong}}^{\text{ref}}(\mathbf{y}|\mathbf{x})} \right]^{\alpha} \right) \right) \\
&= \beta \left( \log\frac{C_A}{C_B} + \alpha \log\frac{\pi_{\text{strong}}^*(\mathbf{y}|\mathbf{x})}{\pi_{\text{strong}}^{\text{ref}}(\mathbf{y}|\mathbf{x})} \right) \\
&= \beta \log\frac{C_A}{C_B} + \alpha \beta \log\frac{\pi_{\text{strong}}^*(\mathbf{y}|\mathbf{x})}{\pi_{\text{strong}}^{\text{ref}}(\mathbf{y}|\mathbf{x})}.
\end{aligned}
\end{equation}
Combining with the definition of the true reward Eq.~\ref{eq:strong-reward} $r(\mathbf{x}, \mathbf{y}) = \beta\log\frac{\pi_{\text{strong}}^*(\mathbf{y}|\mathbf{x})}{\pi_{\text{strong}}^{\text{ref}}(\mathbf{y}|\mathbf{x})}$, we can see:
\begin{equation}
\begin{aligned}
& \quad R_{\text{weak}}(\mathbf{x}, \mathbf{y})   \\
&= \alpha \cdot \left( \beta \log\frac{\pi_{\text{strong}}^*(\mathbf{y}|\mathbf{x})}{\pi_{\text{strong}}^{\text{ref}}(\mathbf{y}|\mathbf{x})} \right) + \beta \log\frac{C_A}{C_B} \\
&= \alpha \cdot r(\mathbf{x}, \mathbf{y}) + \text{Constant}.
\end{aligned}
\end{equation}
Here, $\text{Constant} = \beta \log\frac{C_A}{C_B}$ is an additive constant independent of the specific sequence $\mathbf{y}$.

Therefore, under the two assumptions, the weak model's proxy reward $R_{\text{weak}}(\mathbf{x}, \mathbf{y})$ is proportional to the strong model's true sequence-level alignment reward $r(\mathbf{x}, \mathbf{y})$, plus an additive constant.
This proportional relationship (including an additive constant) has critical implications for optimization: monotonicity. Since the proportionality constant $\alpha$ is positive, a strict monotonic relationship is maintained between the two rewards. This means if $r(\mathbf{x},\mathbf{y}_1) > r(\mathbf{x},\mathbf{y}_2)$, then it must be that $R_{\text{weak}}(\mathbf{x},\mathbf{y}_1) > R_{\text{weak}}(\mathbf{x},\mathbf{y}_2)$, and vice versa. As a result, if our W2-AlignTree's MCTS or re-ranking process aims to maximize $R_{\text{weak}}$, the resulting optimal sequence will also be the optimal sequence that maximizes the true alignment reward $r(\mathbf{x},\mathbf{y})$.

In a practical setting, weak and strong models may not perfectly satisfy the above assumptions. To address this, we can introduce an error term, representing the deviations of the aligned and unaligned weak models from the ideal power-law relationship. In this case, the proxy reward becomes:
\begin{equation}
R_{\text{weak}}(\mathbf{x},\mathbf{y}) = \alpha \cdot r(\mathbf{x},\mathbf{y}) + \Delta'(\mathbf{x},\mathbf{y}),
\end{equation}
where $\Delta'(\mathbf{x},\mathbf{y})$ is an error term encompassing all deviations. As long as the error term $\Delta'(\mathbf{x},\mathbf{y})$ remains small or relatively stable within the set of candidates, it will not change the monotonic ordering of the sequences. Thus, we can still ensure that the selected sequence is close to optimal.

In conclusion, this analysis demonstrates that, under more concise and reasonable assumptions, the weak model's proxy reward is proportional to the strong model's true alignment reward (up to an additive constant) and exhibits a strict monotonic relationship. Even with finite deviations, as long as the error terms are controlled, the proxy reward can effectively guide the search process, ensuring that the final sequence performs excellently with respect to the true alignment objective. This provides a solid theoretical foundation for using a weak model's signal as a proxy within the W2S-AlignTree framework.

\begin{algorithm*}[!ht]
\caption{W2S-AlignTree: Weak-to-Strong Alignment with Monte Carlo Tree Search}
\label{alg:w2s_aligntree}
\begin{algorithmic}[1]
\REQUIRE \textbf{Models and Data}: Input prompt $\mathbf{x}$, Unaligned strong model $\pi_{\text{strong}}$, Aligned weak model $\pi^{*}_{\text{weak}}$, Unaligned weak model $\pi_{\text{weak}}^{\text{ref}}$
\REQUIRE \textbf{Hyperparameters}: MCTS iterations $m$, Step chunk length $L$, Number of expansion candidates $K$, EA-PUCT exploration constant $c$, Entropy-aware weight $w$, Number of top penultimate nodes with children for re-ranking $M$
\ENSURE Best aligned response $\mathbf{y}_{\text{best}}$

\STATE \textbf{Stage 1: Generative Search Tree Construction}
\STATE Initialize a search tree with a root node $s_{\text{root}}$ corresponding to the prompt $\mathbf{x}$.
\FOR{$i=1$ to $m$}

  \STATE // \textbf{Selection}: Traverse from the root to a leaf node using the EA-PUCT strategy.
  \STATE $s_{\text{current}} \leftarrow s_{\text{root}}$
  \WHILE{$s_{\text{current}}$ is not a leaf node}
      \STATE Select child node $s_{\text{child}}$ using the EA-PUCT score:
      \STATE \quad $s_{\text{child}} \leftarrow \arg\max_{s' \in \text{children}(s_{\text{current}})} \left( R(s') + c \cdot P(s') \cdot \frac{\sqrt{N(s_{\text{current}})}}{1 + N(s')} \cdot (1 + w \cdot H(s')) \right)$
      \STATE $s_{\text{current}} \leftarrow s_{\text{child}}$
  \ENDWHILE
  \STATE $s_{\text{leaf}} \leftarrow s_{\text{current}}$

  \STATE // \textbf{Expansion}: Expand the selected leaf node $s_{\text{leaf}}$.
  \STATE Let $\mathbf{y}'$ be the partial prefix (token sequence) corresponding to $s_{\text{leaf}}$.
  \STATE Sample $K$ token sequences, each of length $L$, using the strong model $\pi_{\text{strong}}$ from the state corresponding to $s_{\text{leaf}}$.
  \STATE $Y_{\text{chunks}} \leftarrow \{ \text{sampled\_chunk}_1, \dots, \text{sampled\_chunk}_K \}$
  \FOR{each new token chunk $y_{1:L} \in Y_{\text{chunks}}$}
    \STATE Create a new child node $s'$ for $y_{1:L}$ under $s_{\text{leaf}}$.
    \STATE Let $\mathbf{y}_{\text{new}} \leftarrow \mathbf{y}' \circ y_{1:L}$ // Concatenate the prefix with the new chunk
    \STATE // \textbf{Evaluation}: Compute the proxy reward for the new leaf node.
    \STATE $R(s') \leftarrow \log(\pi^{*}_{\text{weak}}(\mathbf{y}_{\text{new}}|\mathbf{x})) - \log(\pi_{\text{weak}}^{\text{ref}}(\mathbf{y}_{\text{new}}|\mathbf{x}))$
    \STATE If $s'$ meets a stopping condition (e.g., reaching maximum length or generating \verb|EOS| token), set $R(s') \leftarrow -\infty$.
  \ENDFOR

  \STATE // \textbf{Backpropagation}: Update path statistics from new nodes to the root.
  \STATE For each node $s$ on the path from $s_{\text{leaf}}$ to $s_{\text{root}}$:
  \STATE \quad $N(s) \leftarrow N(s) + 1$
  \STATE \quad $R(s) \leftarrow \max(R(s), \max_{s' \in \text{children}(s)} R(s'))$ // Propagate maximum return
\ENDFOR

\STATE \textbf{Stage 2: Best Candidate Decision}
\STATE $S_{\text{penultimate}} \leftarrow \text{find all penultimate nodes in the tree}$
\IF{$S_{\text{penultimate}}$ is empty}
  \STATE // Fallback mechanism: return the sequence corresponding to the node with the highest MCTS return.
  \STATE $\mathbf{y}_{\text{best}} \leftarrow \text{sequence corresponding to } \arg\max_{s \in \text{nodes}(s_{\text{root}})} R(s)$
\ELSE
  \STATE // Select the Top-M penultimate nodes with the highest MCTS return.
  \STATE $S_{\text{top\_M}} \leftarrow \text{select Top-M nodes from } S_{\text{penultimate}} \text{ based on } R(s)$
  \STATE // Collect all children of these Top-M nodes as final candidates.
  \STATE $Y_{\text{candidates}} \leftarrow \text{collect all child sequences of all nodes in } S_{\text{top\_M}}$
  \STATE // Re-rank $M \cdot K$ candidates using the global alignment score.
  \FOR{each candidate sequence $\mathbf{y} \in Y_{\text{candidates}}$}
    \STATE Compute global alignment score: $r(\mathbf{x}, \mathbf{y}) = \log(\pi^{*}_{\text{weak}}(\mathbf{y}|\mathbf{x})) - \log(\pi_{\text{weak}}^{\text{ref}}(\mathbf{y}|\mathbf{x}))$
  \ENDFOR
  \STATE $\mathbf{y}_{\text{best}} \leftarrow \arg\max_{\mathbf{y} \in Y_{\text{candidates}}} r(\mathbf{x}, \mathbf{y})$
\ENDIF

\RETURN $\mathbf{y}_{\text{best}}$
\end{algorithmic}
\end{algorithm*}

\subsection{Algorithm Details}
\subsubsection{A.5.1 Algorithm Workflow and Pseudocode.}
The W2S-AlignTree framework, as detailed in Algorithm \ref{alg:w2s_aligntree}, formalizes LLM alignment as an optimal heuristic search problem during the inference phase. It employs a dual-stage strategy to achieve fine-grained, inference-time alignment without modifying the strong model's parameters.

\textbf{Stage 1: Generative Search Tree Construction}

This stage iteratively builds a generative search tree over $m$ MCTS iterations (Algorithm \ref{alg:w2s_aligntree}, Lines 3-27). Each node in the tree represents a partial token sequence (prefix), and edges correspond to generated token chunks. The process consists of four phases:

\begin{itemize}
\item \textbf{Selection} (Lines 4-11): The algorithm traverses the tree from the root to a leaf node using the EA-PUCT strategy. The EA-PUCT score balances the node's current reward ($R(s')$), its prior probability ($P(s')$), the number of visits to its parent ($N(s_{\text{current}})$) and itself ($N(s')$), and an entropy-aware bonus ($1 + w \cdot H(s')$). This dynamic adjustment helps in balancing exploration and exploitation, adapting to the uncertainty of the strong model's output probability distribution.

\item \textbf{Expansion} (Lines 12-22): Once a leaf node $s_{\text{leaf}}$ is selected, representing a partial prefix ($\mathbf{y}'$), the strong model ($\pi_{\text{strong}}$) is used to sample $K$ new token chunks, each of length $L$. These chunks, denoted as $y_{1:L}$, are then concatenated with the partial prefix $\mathbf{y}'$ to form new sequences $\mathbf{y}_{\text{new}} = \mathbf{y}' \circ y_{1:L}$. For each new chunk, a new child node $s'$ is created under $s_{\text{leaf}}$.

\item \textbf{Evaluation} (Lines 19-21): For each newly created child node $s'$, a proxy reward $R(s')$ is computed. This proxy reward is derived from the Weak-to-Strong Generalization paradigm, using the log-likelihood ratio between an aligned weak model ($\pi^{*}_{\text{weak}}$) and an unaligned weak model ($\pi_{\text{weak}}^{\text{ref}}$). Specifically, $R(s') = \log(\pi^{*}_{\text{weak}}(\mathbf{y}_{\text{new}}|\mathbf{x})) - \log(\pi_{\text{weak}}^{\text{ref}}(\mathbf{y}_{\text{new}}|\mathbf{x}))$. If a node reaches a stopping condition (e.g., reaching maximum length or generating the \texttt{EOS} token), its reward is set to $-\infty$ to prevent further expansion, but it remains a candidate for final evaluation.

\item \textbf{Backpropagation} (Lines 23-26): The reward $R(s')$ obtained from the newly evaluated node is back-propagated up the tree to the root. For each node $s$ on the path from $s_{\text{leaf}}$ to $s_{\text{root}}$, its visit count $N(s)$ is incremented, and its accumulated reward $R(s)$ is updated to be the maximum of its current value and the maximum reward of its children. This "maximum return propagation" ensures that high-value trajectories are preserved and computational resources are focused on the most promising paths.
\end{itemize}

\textbf{Stage 2: Best Candidate Decision}

After the MCTS iterations are complete, the second stage focuses on identifying the best-aligned complete response from the constructed generative search tree (Algorithm \ref{alg:w2s_aligntree}, Lines 29-41).

\begin{itemize}
\item \textbf{Candidate Collection} (Lines 30-36): The algorithm first identifies all ``penultimate nodes" in the tree. These are nodes that are one step away from forming a complete sequence. If no penultimate nodes are found (e.g., due to insufficient search budget), a fallback mechanism selects the sequence corresponding to the node with the highest MCTS return across the entire tree. Otherwise, the algorithm selects the Top-M penultimate nodes with the highest MCTS returns. All child sequences stemming from these Top-M penultimate nodes are then collected into a set $Y_{\text{candidates}}$. This ensures that a diverse set of high-potential complete responses are considered.

\item \textbf{Global Re-ranking} (Lines 37-41): Each candidate sequence $\mathbf{y} \in Y_{\text{candidates}}$ is then re-ranked using a global alignment score, $r(\mathbf{x}, \mathbf{y}) = \log(\pi^{*}_{\text{weak}}(\mathbf{y}|\mathbf{x})) - \log(\pi_{\text{weak}}^{\text{ref}}(\mathbf{y}|\mathbf{x}))$. This score represents the full sequence-level alignment quality. The candidate sequence with the highest global alignment score is ultimately selected as the best aligned response $\mathbf{y}_{\text{best}}$.
\end{itemize}

This dual-stage approach combines the efficient exploration capabilities of MCTS with a robust global re-ranking mechanism, guided by weak model signals, to achieve effective inference-time alignment for LLMs.

\subsubsection{A.5.2 More Interpretation of EA-PUCT.}

In the context of the W2S-AlignTree framework, the advantages of EA-PUCT over traditional UCT algorithms are primarily manifested in its dynamic management of the exploration-exploitation trade-off, uncertainty modeling, and effective synergy with weak supervision signals. Unlike the traditional UCT selection strategy, which relies on a static constant $c$ to balance exploration and exploitation, EA-PUCT introduces a dynamic adjustment mechanism based on Shannon entropy, leading to a more intelligent and efficient tree search strategy.

The traditional UCT selection strategy is based on the following formula:
\begin{equation}
\label{eq:uct}
\text{UCT}(s) = \bar{Q}(s) + c \sqrt{\frac{\ln N(s_\text{p})}{N(s)}}.
\end{equation}
Here, $\bar{Q}(s)$ is the average historical return for node $s$, $N(s)$ is the number of visits, and $s_p$ is the parent node. This static strategy may lead to premature convergence in high-confidence regions, causing the algorithm to overlook potentially high-value but low-frequency nodes.

EA-PUCT fundamentally improves upon this strategy by introducing a dynamic weight associated with the Shannon entropy $H(s)$ of state $s$. The selection strategy, known as EUU (Entropy-Uncertainty Upper Confidence), can be formalized as:
\begin{equation}
{E\text{-}PU}(s) = R(s) + c \cdot P(s) \cdot \frac{\sqrt{N(s_{\text{p}})}}{1+N(s)} \cdot (1 + w \cdot H(s)).
\end{equation}
The core improvement lies in the dynamic weight term $(1 + w \cdot H(s))$. The entropy $H(s) = -\sum_a P(s,a) \log P(s,a)$ quantifies the model's cognitive uncertainty in state $s$. When the model is highly confident in a certain branch (i.e., low $H(s)$), this weight approaches 1, suppressing unnecessary exploration and reinforcing the exploitation of known high-value paths. Conversely, when the model faces high uncertainty (i.e., high $H(s)$), the weight increases significantly, actively elevating the priority of exploration to avoid getting trapped in local optima. This adaptive exploration mechanism is theoretically closer to dynamic strategies for optimal stopping problems (such as the Gittins index) but is computationally more efficient.

From an information-theoretic perspective, incorporating entropy as an exploration bonus is equivalent to introducing an information-theoretic regularizer into the policy optimization objective. The objective function of EA-PUCT can be understood as maximizing a combination of return and entropy reward:
\begin{equation}
\label{eq:max_entropy_rl}
\max_{\pi} \mathbb{E}_{\pi} \left[ R(s) + \lambda \cdot H(\pi(\cdot|s)) \right].
\end{equation}
This objective aligns with the core idea of Maximum Entropy Reinforcement Learning (MaxEnt RL), but it is applied to tree search rather than parameterized policies. Its theoretical significance is twofold: first, by encouraging the exploration of high-entropy actions, EA-PUCT effectively prevents policy collapse, a phenomenon where MCTS gets trapped in a single high-probability path due to greedy selection. Second, by rewarding diversity, it encourages the exploration of low-probability but potentially high-value actions, which is highly consistent with the principle of Optimism in the Face of Uncertainty (OFU).

Furthermore, EA-PUCT's entropy-aware mechanism exhibits a crucial theoretical synergy with the weak supervision signals in W2SG. The proxy value $V_{\text{proxy}}$ provided by the weak model $\pi_{\text{weak}}^*$ may contain systematic biases. High-entropy regions serve as a clear signal to the strong model $\pi_{\text{strong}}$ that the weak supervision is potentially unreliable in these areas, necessitating autonomous exploration to correct its limitations. Analogous to Bayesian posterior sampling, EA-PUCT uses entropy as an indicator of ``prior uncertainty," guiding the strong model to increase exploration in regions where the weak supervision's confidence is low, thereby achieving a gradual alignment of preferences.

Finally, regarding computational complexity and convergence, EA-PUCT also presents advantages. The entropy calculation depends only on the current policy distribution, with a computational overhead of $O(|A|)$ (where $|A|$ is the size of the action space), which is negligible compared to the simulation phase of MCTS. Although the introduction of the entropy term adds a certain bias, it can be proven that by appropriately selecting $w$, the algorithm still satisfies asymptotic optimality; that is, as the number of searches $N \to \infty$, the selection probability converges to the optimal action. The proof follows a similar line of reasoning to the finite-time bounds of UCB algorithms but requires additional control over the variance of the entropy term.

\subsubsection{A.5.3 Complexity and Resource Analysis.}

While W2S-AlignTree demonstrates superior alignment performance, it is important to analyze the computational and resource it introduces. The complexity of our framework primarily stems from two core components: the generative MCTS process and the dual-model (weak/strong) evaluation architecture.

\paragraph{Computational Complexity.}
The computational complexity of W2S-AlignTree is dominated by the MCTS loop. A single MCTS iteration consists of four main steps: selection, expansion, evaluation, and backpropagation. The most resource-intensive steps are evaluation and expansion, which require forward passes through the models.

Assuming $m$ MCTS iterations, an expansion branching factor of $K$, and a step chunk length of $L$, the computational complexity can be approximated as follows:
\begin{itemize}
    \item \textbf{LLM Forward Passes}: Each MCTS iteration requires one strong LLM forward pass to sample $K$ new token chunks, and $K$ weak LLM forward passes to evaluate these new chunks. This results in a total of $m \times (1+K)$ forward passes.
    \item \textbf{Trade-off between Strong and Weak LLMs}: Critically, the computational cost is effectively distributed. Each iteration involves only one forward pass through the large strong LLM ($\pi_{\text{strong}}$), while the evaluation of the $K$ new nodes relies on the significantly smaller and faster weak LLM ($\pi_{\text{weak}}$). Consequently, the total computational time per MCTS iteration is only a small constant multiple of a single strong LLM forward pass, plus the cost of $K$ fast weak LLM evaluations. This approach is more efficient than other methods that require multiple runs of the strong LLM (e.g., re-ranking over full sequences).
\end{itemize}
The total complexity can be expressed as $\mathcal{O}(m \cdot (T_{\text{strong}} + K \cdot T_{\text{weak}}))$, where $T_{\text{strong}}$ and $T_{\text{weak}}$ are the time complexities of a single forward pass for the strong and weak LLMs, respectively. Since $T_{\text{weak}} \ll T_{\text{strong}}$, the total complexity is roughly proportional to $m \cdot T_{\text{strong}}$, which is typically far lower than the cost of a brute-force search over all possible sequences.

\paragraph{Memory and Resource Consumption.}
The memory footprint of the W2S-AlignTree consists of two main parts:
\begin{itemize}
    \item \textbf{Model Parameters}: The framework requires loading both the strong and weak LLMs into memory simultaneously. This dual-model architecture necessitates a larger GPU memory footprint compared to single-model inference methods. However, this overhead is effectively mitigated by choosing a weak LLM that is significantly smaller than the strong LLM.
    \item \textbf{Search Tree}: The MCTS search tree itself consumes memory. The size of the tree is bounded by the product of the number of MCTS iterations $m$ and the expansion branching factor $K$. The memory required for each node stores the partial prefix, statistics (e.g., return $R$, visit count $N$, prior probability $P$), and child pointers. For typical MCTS configurations, the total tree memory grows linearly with $m \times K$, which is manageable.
\end{itemize}
To further alleviate the dual-model memory overhead, strategies such as model quantization or offloading one of the models to CPU memory can be employed, though at the cost of increased latency.

\paragraph{Comparison with Baselines.}
W2S-AlignTree is a novel alignment paradigm, and its advantages and limitations are highlighted in a comparison with different baselines. We compare it with traditional inference-time decoding methods and with mainstream training-time alignment methods.
\begin{itemize}
    \item \textbf{Comparison with Traditional Decoding Methods}: Compared to standard decoding methods like greedy search or beam search, W2S-AlignTree is indeed slower in terms of inference latency. However, this overhead is a deliberate trade-off for achieving higher alignment quality through an intelligent search process during generation. Our experimental results show this trade-off is justified, as it leads to significant performance improvements. Furthermore, by using a considerably smaller and faster weak LLM as a proxy reward function, our search process is more efficient than if we were to use the strong LLM for all evaluations.
    \item \textbf{Comparison with Training-time Alignment Methods}: W2S-AlignTree presents distinct advantages over mainstream training-time alignment methods such as PPO and DPO. Both PPO and DPO are training-time methods that require large amounts of preference data or reward signals to fine-tune the strong LLM's parameters. This process is computationally expensive and highly data-dependent. In contrast, W2S-AlignTree is an \textbf{inference-time alignment} alternative. It does not modify the strong LLM's parameters but instead searches for the optimal aligned sequence during the generation phase via MCTS. This gives W2S-AlignTree key advantages. For example, it avoids the need for expensive parameter fine-tuning and large-scale computational resources, making it more appealing for new tasks or resource-constrained scenarios. Simultaneously, it leverages the idea of ``weak-to-strong generalization" by using a smaller, aligned weak LLM as a proxy reward function. This weak LLM can be trained on relatively easy-to-obtain data, which significantly reduces the reliance on expensive human-annotated data for the alignment process.
\end{itemize}
In summary, W2S-AlignTree offers an alignment paradigm that is complementary to training-time methods like PPO and DPO. By performing MCTS search at inference time and utilizing a weak LLM as a proxy reward, it effectively addresses the dependency on large-scale computational resources and expensive preference data. This makes W2S-AlignTree a more cost-effective and flexible alignment approach, particularly suitable for resource-constrained scenarios or those requiring rapid adaptation to new tasks.

\section{Further Details on the Experimental Setup}
\subsection{Sentiment Generation}
\subsubsection{B.1.1 Model Specification}
Table 1 lists the models used in both the controlled-sentiment generation and summarization tasks, along with their corresponding Hugging Face links.

\begin{table}[!ht]
\centering
\scriptsize
\begin{tabular}{@{}l@{\hskip 4pt}l@{}}
\toprule
\textbf{Models} & \textbf{Links} \\
\midrule
\multicolumn{2}{c}{\textbf{Inference Strong Models}} \\
\midrule
gpt2 & \url{https://huggingface.co/openai-community/gpt2} \\
gpt2-large & \url{https://huggingface.co/openai-community/gpt2-large} \\
gpt2-xl & \url{https://huggingface.co/openai-community/gpt2-xl} \\
Qwen2.5-7B & \url{https://huggingface.co/Qwen/Qwen2.5-7B} \\
Qwen2.5-7B Instruct & \url{https://huggingface.co/Qwen/Qwen2.5-7B-Instruct} \\
Llama-2-7b-hf & \url{https://huggingface.co/meta-llama/Llama-2-7b-hf} \\
Llama-2-7b-chat-hf & \url{https://huggingface.co/meta-llama/Llama-2-7b-chat-hf} \\
Llama-3-8B & \url{https://huggingface.co/meta-llama/Meta-Llama-3-8B} \\
Llama-3-8B-Instruct & \url{https://huggingface.co/meta-llama/Meta-Llama-3-8B-Instruct} \\
tulu-2-7b & \url{https://huggingface.co/allenai/tulu-2-7b} \\
tulu-2-dpo-7b & \url{https://huggingface.co/allenai/tulu-2-dpo-7b} \\
\midrule
\multicolumn{2}{c}{\textbf{Open-source Guidance Weak Models}} \\
\midrule
gpt2-imdb & \url{https://huggingface.co/lvwerra/gpt2-imdb} \\
Llama-3.2-1B & \url{https://huggingface.co/meta-llama/Llama-3.2-1B} \\
Llama-3.2-1B-Instruct & \url{https://huggingface.co/meta-llama/Llama-3.2-1B-Instruct} \\
Qwen2.5-0.5B & \url{https://huggingface.co/Qwen/Qwen2.5-0.5B} \\
Qwen2.5-0.5B-Instruct & \url{https://huggingface.co/Qwen/Qwen2.5-0.5B-Instruct} \\
\bottomrule
\end{tabular}
\caption{Model Specification cross Tasks.}
\label{tab:sentiment_model_spec}
\end{table}

\subsubsection{B.1.2 Hyperparameters for Sentiment Generation}
We frame this task as a controlled generation task for continuing fixed-length film emotional content, with the length of the newly generated content fixed at 50.
We determine the optimal hyperparameters for each method through empirical comparison. 
For all methods, we adopt temperature $T = 0.7$, top-$k = 50$, and top-$p = 1.0$ when sampling from the strong LLMs. 
For CBS, we use $W, K, L = 4, 4, 5$ ($W$: beam width, $K$: successors per state, $L$: chunk length) and for BoN, we use $N = 16$. For our W2S-AlignTree, we set $L = 1$ based on preliminary validation, and report the best results among $c \in \{1.0, 1.5, 2.0\}$.

\subsubsection{B.1.3 Gold Reward Model Details}
We construct a synthetic preference setup where gold reward models approximate human judgments by producing binary preference labels~\cite{gao2023scaling, lightman2023let, rafailov2023direct, zhou2024weak}.
We utilize the publicly available \texttt{distilbert-imdb} model to instantiate the gold reward function $r_{\text{gold}}$. This model is a fine-tuned sentiment classifier $p$ trained on the \texttt{IMDB} dataset~\cite{maas2011learning}. The reward model achieves a high validation accuracy of 0.93, demonstrating a strong correlation with human judgments. We compute the reward as 
\begin{equation}
    r_{\text{gold}}(x, y) = \log p(\text{positive} | x, y) - \log p(\text{negative} | x, y),
\end{equation}
which encourages outputs with positive sentiment.
To obtain synthetic preferences, we use truncated movie reviews as prompts $x$, and generate pairwise completions $(y_1, y_2)$ using \texttt{gpt2-imdb}. Preferences are then derived by comparing the rewards assigned to each completion:
\begin{equation}
p(y_1 \succ y_2 | x) = \sigma \big(r_{\text{gold}}(x, y_1) - r_{\text{gold}}(x, y_2)\big),
\end{equation}
where $\sigma(\cdot)$ denotes the sigmoid function.

\subsubsection{B.1.4 Training and Evaluation Setup}
Direct tuning on the synthetic preferences $\mathcal{D} = \{(x, y_w, y_l)_i\}_{i=1}^N$ involves two stages: SFT and DPO. 
We perform both stages of SFT and DPO using the \texttt{LLaMA-Factory} framework~\cite{zheng2024llamafactory}, without further modification to its default training pipeline.
For DPO, we set the inverse temperature $\beta = 0.1$ to scale the reward differences when optimizing preferences.
All the training and evaluation procedures, including testing, are conducted on a single NVIDIA A800 GPU.

\subsubsection{B.1.5 Prompt Template for Sampling}
To guide large pre-trained models effectively, task-specific prompts are necessary.
We use the following zero-shot instruction:

\begin{quote}
\small
\texttt{Here is a movie review: \{prompt\}}
\end{quote}

\begin{table}[!ht]
\centering
\scriptsize

\begin{tabular}{@{}l@{\hskip 4pt}l@{}}
\toprule
\multicolumn{2}{c}{\textbf{Gold Reward Models}} \\
\midrule
\textbf{Models} & \textbf{Links} \\
\midrule
distilbert-imdb & \url{https://huggingface.co/openai-community/gpt2} \\
Llama-2-7b-hf & \url{https://huggingface.co/meta-llama/Llama-2-7b} \\
oasst-rm-2-pythia-6.9b & \makecell[l]{\url{https://huggingface.co/OpenAssistant/}\\\url{oasst-rm-2-pythia-6.9b-epoch-1}} \\
UltraRM-13b & \url{https://huggingface.co/openbmb/UltraRM-13b} \\

\midrule
\multicolumn{2}{c}{\textbf{Datasets}} \\
\midrule
\textbf{Datasets} & \textbf{Links} \\
\midrule
IMDB & \url{https://huggingface.co/datasets/ZHZisZZ/imdb_preference} \\
Summarize\_from\_Feedback & \makecell[l]{\url{https://huggingface.co/datasets/}\\\url{OpenAssistant/summarize_from_feedback}} \\
TL;DR & \url{https://huggingface.co/datasets/trl-lib/tldr} \\
OASST1 & \url{https://huggingface.co/datasets/OpenAssistant/oasst1} \\
\bottomrule
\end{tabular}

\caption{Gold Reward Models and Datasets across Tasks.}
\label{tab:reward_model_dataset_spec}
\end{table}

\begin{table*}[!htbp]
\centering
\ifdefined\colw
\else
    \newlength\colw
\fi

\setlength\colw{0.10\textwidth}
\setlength\tabcolsep{2pt}
\renewcommand\arraystretch{1.05}

\begin{tabular}{l*{8}{>{\centering\arraybackslash}p{\colw}}}
\toprule
  & GPT2-Large & GPT2-XL & Qwen2.5-7B & Qwen2.5-7B-Instruct &
  Llama-2-7b-hf & Llama-2-7b-chat-hf & Llama-3-8B & Llama-3-8B-Instruct
\\ \midrule

\multicolumn{9}{c}{\makecell{\textbf{Sentiment Generation} \\
              GPT-2 DPO / GPT-2 SFT: 4.09 / 1.09}}

\\ \midrule
Base  & 1.95$\pm$0.05 & 1.51$\pm$0.08 & 1.26$\pm$0.09 & 0.31$\pm$0.14 & 2.05$\pm$0.05 & 2.85$\pm$0.05 & 2.25$\pm$0.04 & 2.67$\pm$0.10 \\
BoN   & 4.03$\pm$0.09 & 3.63$\pm$0.04 & 3.95$\pm$0.04 & 4.07$\pm$0.04 & 3.76$\pm$0.07 & 4.43$\pm$0.04 & 3.98$\pm$0.03 & 4.19$\pm$0.03 \\
CBS   & 4.70$\pm$0.08 & 4.35$\pm$0.01 & 4.36$\pm$0.01 & 4.54$\pm$0.01 & 4.48$\pm$0.03 & 4.52$\pm$0.05 & 4.53$\pm$0.06 & 4.45$\pm$0.01 \\
W2S-AT (Ours) & \textbf{4.84}$\pm$0.05 & \textbf{4.50}$\pm$0.01 & \textbf{4.79}$\pm$0.02 & \textbf{4.74}$\pm$0.06 & \textbf{4.50}$\pm$0.01 & \textbf{4.65}$\pm$0.03 & \textbf{4.78}$\pm$0.01 & \textbf{4.74}$\pm$0.03 
\\ \midrule
\multicolumn{9}{c}{\makecell{\textbf{Summarization} \\
              GPT-2 DPO / GPT-2 SFT: 0.12 / -0.25}}
\\ \midrule
Base  & -0.60$\pm$0.03 & -0.08$\pm$0.07 & 1.73$\pm$0.07 & 1.56$\pm$0.05 & 1.20$\pm$0.03 & 2.14$\pm$0.03 & 1.57$\pm$0.05 & 1.39$\pm$0.04 \\
BoN   & 0.33$\pm$0.04  & 0.08$\pm$0.03  & 1.60$\pm$0.03 & 1.95$\pm$0.03 & 1.86$\pm$0.03 & 1.81$\pm$0.01 & 1.43$\pm$0.04 & 2.26$\pm$0.02 \\
CBS   & 0.38$\pm$0.02  & 0.48$\pm$0.02  & 1.66$\pm$0.02 & \textbf{2.08}$\pm$0.02 & 1.83$\pm$0.04 & 1.33$\pm$0.03 & 1.89$\pm$0.03 & 2.42$\pm$0.04 \\
W2S-AT (Ours) & \textbf{0.41}$\pm$0.03 & \textbf{0.84}$\pm$0.04 & \textbf{2.03}$\pm$0.01 & 2.02$\pm$0.02 & \textbf{2.00}$\pm$0.01 & \textbf{2.78}$\pm$0.02 & \textbf{2.19}$\pm$0.01 & \textbf{2.46}$\pm$0.03 \\

\bottomrule
\end{tabular}
\caption{
Evaluation of alignment performance on sentiment generation and summarization, comparing W2S-AlignTree (W2S-AT) with representative baseline methods.
We use \texttt{GPT-2 SFT} and \texttt{GPT-2 DPO} as weak guidance models,
\textbf{Bold} indicates the best $r_{\mathrm{gold}}$ score.
The scores for sentiment generation and summarization are derived from the reward model \texttt{distilbert-imdb} and a custom reward model fine-tuned on \texttt{Llama-2-7b-hf}, respectively.
}
\label{tab:gpt2_number}
\end{table*}


\subsection{Summarization}
\subsubsection*{B.2.1 Model Specification}

The summarization task employs the same set of models as used in the Sentiment Generation task.  
For details and corresponding Hugging Face links, refer to Table~\ref{tab:sentiment_model_spec}.

\subsubsection{B.2.2 Hyperparameters for Summarization}

For the summarization task, we adopt the same decoding and sampling hyperparameters as used in the controlled-sentiment generation task.  
For W2S-AlignTree, we report the best results with $L \in \{3, 4, 5\}$ and $c \in \{1.0, 1.5, 2.0\}$ based on validation performance.

\subsubsection{B.2.3 Gold Reward Model Details}
We build the gold reward model $r_{\text{gold}}$ for the summarization task by fine-tuning \texttt{llama-2-7b-hf} with LoRA method~\cite{hu2022lora} on the \texttt{summarize\_from\_feedback} dataset~\cite{stiennon2020learning} and \texttt{LLaMA-Factory} framework. 
The training setup includes a linear projection head and binary cross-entropy loss, with a batch size of 16, a learning rate of $1\mathrm{e}{-5}$ for the projection head and $5\mathrm{e}{-6}$ for the remaining parameters. 
Training is performed over one epoch with a cosine learning rate schedule. 
The reward model achieves a validation accuracy of 0.72. Given this imperfect performance, we introduce a relabeling strategy in the subsequent stages to further ensure the accuracy of the evaluation.

Synthetic preferences are derived by comparing reward scores assigned to pairwise responses:
\begin{equation}
p(\mathbf{y}_1 \succ \mathbf{y}_2 \mid \mathbf{x}) = \sigma \big( r_{\text{gold}}(\mathbf{x}, \mathbf{y}_1) - r_{\text{gold}}(\mathbf{x}, \mathbf{y}_2) \big),
\end{equation}
where $\sigma(\cdot)$ denotes the sigmoid function.

\subsubsection{B.2.4 Training and Evaluation Setup}

Direct tuning on the synthetic preferences $\mathcal{D} = \{(x, y_w, y_l)_i\}_{i=1}^N$ for the summarization task follows the same two-stage process: SFT and DPO. 
To obtain high-quality preferences, we relabel the original data using the summarization-specific gold reward model described in Section~B.2.3.
The relabeled dataset better captures reward-consistent supervision, resulting in improved learning signals during SFT and DPO. For DPO, we set the inverse temperature $\beta = 0.1$ to calibrate the reward differences when optimizing preferences.
The \texttt{TL;DR}~\cite{2019TLDR} dataset rather than \texttt{summarize\_from\_feedback} is selected to evaluate the model's robustness and generalization capabilities.
All training and evaluation procedures, including testing, are conducted on a single NVIDIA A800 GPU.

\subsubsection{B.2.5 Prompt Template for Sampling}
To guide large pre-trained models effectively, task-specific prompts are necessary. For the summarization task, we use the following zero-shot format~\cite{2019TLDR}:

\begin{quote}
\small
\texttt{SUBREDDIT: r/\{subreddit\}} \\
\texttt{TITLE: \{title\}} \\
\texttt{POST: \{post\}} \\
\texttt{TL;DR:}
\end{quote}

\subsection{Instruction-Following}
\subsubsection*{B.3.1 Model Specification}
The models used in the instruction-following task are included in Table~\ref{tab:sentiment_model_spec}. 
In particular, we additionally report results on \texttt{Tulu-2-7b} and \texttt{Tulu-2-dpo-7b}~\cite{ivison2023camels}.

\subsubsection{B.3.2 Gold Reward Model Details}
We choose the \texttt{OASST1}~\cite{kopf2023openassistant} dataset for further assessment.
To evaluate instruction-following capabilities, we adopt two distinct reward models as gold evaluators:
\begin{itemize}
    \item \textbf{\texttt{oasst-rm-2-pythia-6.9b}}~\cite{OpenAssistantOASSTRM2}: 
    this is a task-aligned reward model specifically fine-tuned on the \texttt{OASST1} dataset to reflect the instruction-specific alignment quality.
    \item \textbf{\texttt{UltraRM-13b}}~\cite{cui2023ultrafeedback}: 
    this is a general-purpose reward model designed for instruction evaluation across diverse domains, exhibiting broad generalization capabilities.
\end{itemize}
We use the reward score $r_{\text{gold}}$ produced by these models to assess the quality of model output. Higher reward scores indicate better alignment with human-preferred responses.

\subsubsection*{B.3.3 \quad Compute Resources Specification}
Model inference and evaluation for the Instruction-Following task are conducted on a single NVIDIA A800 GPU.

\section{Additional Experimental Results}

\subsection{Sentiment Generation \& Summarization}
To complement the bar plots presented in the main text, we provide detailed numerical results for the sentiment generation and summarization tasks in Table~\ref{tab:gpt2_number}. This table reports the $r_{\mathrm{gold}}$ scores obtained by different alignment methods across a diverse set of models. Results are shown for both tasks using distinct weak guidance models (GPT-2 SFT and GPT-2 DPO), both of which are trained by us, enabling a fair and consistent comparison. Bold entries indicate the best-performing method under each configuration.
Based on the bar charts in the main text and the detailed numerical results in Table~\ref{tab:gpt2_number}, the W2S-AlignTree method achieves significant alignment results in both sentiment generation and summarization tasks. Across all tested configurations, its $r_{\mathrm{gold}}$ scores are generally superior to those of all baseline methods, demonstrating the superiority of W2S-AlignTree.

\begin{table}[!htbp]
\centering
\begin{tabular}{lcc}
\toprule
\textbf{Models} & \textbf{W2S-AlignTree} & \textbf{DPO} \\
\midrule
GPT2-XL        & 4.50 & 4.86 \\
GPT2-Large     & 4.84 & 5.01 \\
Llama-3-8B     & 4.78 & 4.82 \\
Llama-2-7b-hf  & 4.50 & 4.43 \\
Qwen2.5-7B     & 4.79 & 4.88 \\
\bottomrule
\end{tabular}
\caption{
Comparison of \texttt{W2S-AlignTree} and \texttt{DPO} on different backbone models using Gold RM scores.
}
\label{tab:w2s_vs_dpo}
\end{table}

To evaluate the overall alignment quality, we further compare W2S-AlignTree with DPO-fine-tuned models across a diverse set of LLMs, as summarized in Table~\ref{tab:w2s_vs_dpo}. 
Although W2S-AlignTree exhibits slightly lower scores in some cases, it remains highly competitive without requiring any supervised fine-tuning, demonstrating its effectiveness as a plug-and-play alignment method.

\begin{table*}[!htbp]
\centering
\setlength\colw{0.10\textwidth}
\setlength\tabcolsep{2pt}
\renewcommand\arraystretch{0.96}

\begin{tabular}{l*{8}{>{\centering\arraybackslash}p{\colw}}}
\toprule
& Qwen2.5-7B & Qwen2.5-7B-Instruct & Llama3-8B & Llama3-8B-Instruct &
  Llama2-7b-hf & Llama2-7b-chat-hf & tulu2-7b & tulu2-7b-dpo \\ \midrule

\multicolumn{9}{c}{\makecell{\textbf{Gold Reward Model: oasst-rm-2-pythia-6.9b} \\
                          Qwen2.5-0.5B-Instruct / Qwen2.5-0.5B: 0.58 / 0.35}}
\\ \midrule
Greedy  & 0.80 & 1.45 & -0.68 & 0.71 & -0.75 & 0.79 & -0.13 & 0.52 \\
BoN     & 0.81 & 1.20 & -0.65 & 0.73 & -0.95 & \textbf{1.14} & 0.36 & 0.53 \\
CBS     & 0.74 & \textbf{1.49} & -1.31 & 0.69 & -0.83 & 0.44 & -1.28 & -0.47 \\
W2S-AT (Ours) & \textbf{0.90} & 1.14 & \textbf{-0.55} & \textbf{0.74} & \textbf{-0.70} & 0.73 & \textbf{0.46} & \textbf{0.54} \\ \midrule

\multicolumn{9}{c}{\makecell{\textbf{Gold Reward Model: UltraRM-13b} \\
                          Qwen2.5-0.5B-Instruct / Qwen2.5-0.5B: -5.73 / -7.36}}
\\ \midrule
Greedy  & -3.15 & 1.12 & -9.15 & -3.45 & -9.56 & -4.29 & -6.48 & -5.80 \\
BoN     & -2.97 & -0.04 & -8.95 & -3.37 & -9.72 & -4.24 & -5.88 & -5.16 \\
CBS     & -3.05 & \textbf{1.40} & -10.30 & -3.38 & -8.96 & -6.88 & -9.94 & -7.74 \\
W2S-AT (Ours) & \textbf{-2.88} & 1.04 & \textbf{-7.56} & \textbf{-3.28} & \textbf{-8.67} & \textbf{-3.81} & \textbf{-5.35} & \textbf{-5.00} \\
\bottomrule
\end{tabular}
\caption{
Evaluating instruction-following on \texttt{OASST1} using W2S-AlignTree (W2S-AT) and representative baselines.
We use \texttt{Qwen2.5-0.5B-Instruct} and \texttt{Qwen2.5-0.5B} as weak guidance models,
and both \texttt{oasst-rm-2-pythia-6.9b} and \texttt{UltraRM-13b} as gold reward models.
\textbf{Bold} indicates the best $r_{\mathrm{gold}}$ score in each column.
}
\label{tab:qwen2.5_0.5b_ins}
\end{table*}

\begin{figure*}[!ht]
  \centering
  \includegraphics[width=0.95\linewidth]{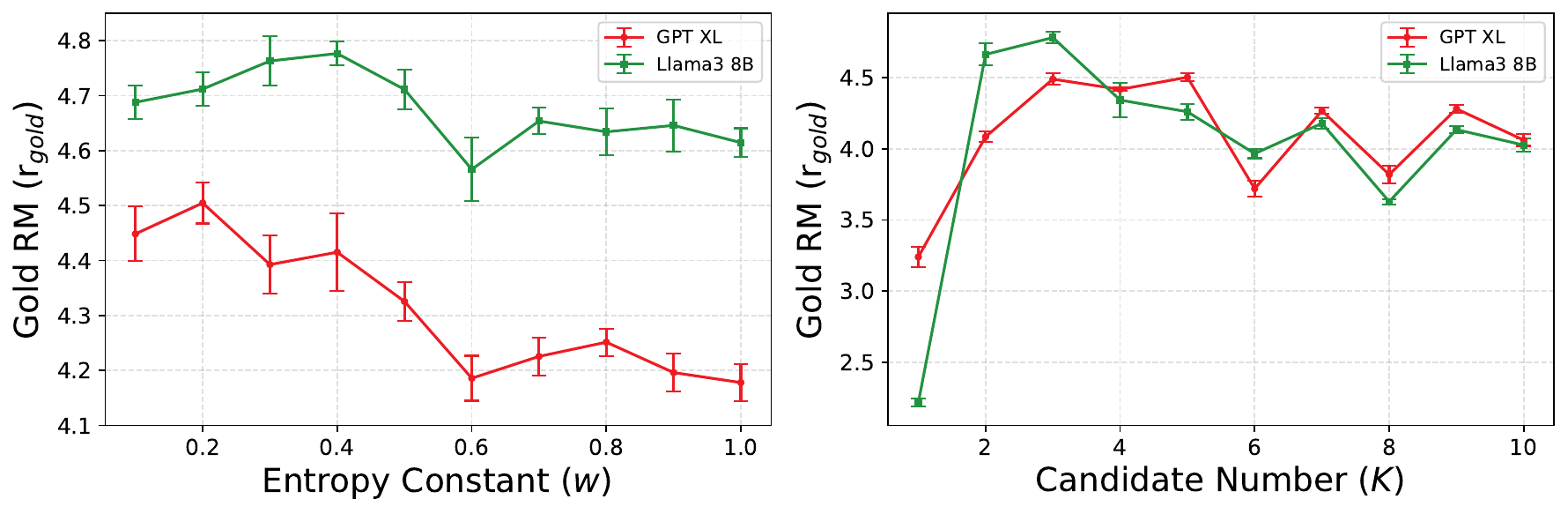}
  \caption{Further analysis of hyperparameter sensitivity with respect to (left) entropy constant \( w \in [0.1, 1.0] \) and (right) candidate number \( K \in [1, 10] \).We conduct evaluations on both \texttt{GPT XL} and \texttt{Llama-3-8B} under varying entropy settings and candidate counts.
  }
  \label{fig:combined_results_wk}
\end{figure*}

\subsection{Instruction-Following}

To assess the robustness and generalizability of W2S-AlignTree with respect to different weak guidance sources, we conducted a cross-validation study. In this study, we followed the same evaluation protocol as in the main text, replacing the weak guidance models from \texttt{Llama-3.2-1B-Instruct/Llama-3.2-1B} with the cross-family and smaller-scale models \texttt{Qwen2.5-0.5B-Instruct} and \texttt{Qwen2.5-0.5B}, while the target models, baselines (Base, BoN, CBS), and gold reward models (\texttt{oasst-rm-2-pythia-6.9b} and \texttt{UltraRM-13b}) remained unchanged. The detailed results are presented in Table~\ref{tab:qwen2.5_0.5b_ins}.

Using the \texttt{oasst-rm-2-pythia-6.9b} evaluator, W2S-AlignTree achieved the best or second-best ($r_{\text{gold}}$) scores on most target models. Our method delivered significant improvements across several models. For example, the score for \texttt{Llama3-8B} improved from $-0.68$ to $-0.55$ (an increase of $0.13$), and the score for \texttt{tulu2-7B} saw a remarkable leap from $-0.13$ to $0.46$ (an increase of $0.58$), demonstrating impressive alignment effectiveness. Even in a few cases, such as \texttt{Qwen2.5-7B-Instruct}, where \texttt{CBS} or \texttt{BoN} slightly outperformed W2S-AlignTree, our method remained highly competitive, achieving a robust performance increase on \texttt{Qwen2.5-7B} from $0.80$ to $0.90$ (an increase of $0.10$).

Under the more stringent \texttt{UltraRM-13b} evaluator, the advantages of W2S-AlignTree were even clearer. Almost all target models showed greater performance improvements. For instance, the score for \texttt{Llama3-8B} jumped from $-9.15$ to $-7.56$ (an increase of $1.59$), \texttt{Llama2-7B-hf} improved from $-9.56$ to $-8.67$ (an increase of $0.89$), and \texttt{tulu2-7B} increased from $-6.48$ to $-5.35$ (an increase of $1.14$). These results indicate that W2S-AlignTree is capable of providing more powerful alignment capabilities when facing more challenging evaluation criteria.

In conclusion, the core finding of instruction-following task is that the effectiveness of W2S-AlignTree does not rely on a particular family or scale of weak guidance models. When the weak guidance source was switched to the \texttt{Qwen} model, which is from a different family and is smaller in scale than the target models like \texttt{Llama} and \texttt{tulu}, W2S-AlignTree consistently and robustly improved scores under both evaluators and across both base and instruct target models. 
These results provide strong evidence for the cross-model transferability and guidance-source diversity of W2S-AlignTree, supporting its potential as a plug-and-play inference-time alignment method.

\subsection{Ablation Studies}
We further analyze the hyperparameter sensitivity of \textit{W2S-AlignTree}, under different inference-time hyperparameter configurations. Specifically, we examine the effects of the \textit{entropy constant} $w \in [0.1, 1.0]$, which controls the degree of exploration during search, and the \textit{candidate number} $K \in [1, 10]$ for MCTS expansion. 

Experiments are conducted on the \textit{sentiment generation} task using both \texttt{GPT-XL} and \texttt{Llama-3-8B}. 
As shown in Fig.~\ref{fig:combined_results_wk} (left), increasing $w$ enhances exploration, but leads to divergent trends across models. 
\texttt{Llama-3-8B} benefits from moderate exploration, reaching peak performance of $w$ from $0.3$ to $0.4$, while excessive entropy ($w > 0.6$) degrades performance. 
In contrast, \texttt{GPT-XL} exhibits better alignment at smaller $w$, suggesting a stronger reliance on exploitative search.
Fig.~\ref{fig:combined_results_wk} (right) shows both models achieve optimal results at moderate candidate numbers $K$ from $3$ to $5$, whereas larger values of $K$ introduce unnecessary search variance and lead to diminishing returns. 

These results indicate that \textit{W2S-AlignTree} effectively leverages model-specific characteristics through entropy-aware exploration and candidate filtering during inference-time alignment.
Furthermore, the stable trends across a wide range of parameter settings demonstrate the robustness and reliability of \textit{W2S-AlignTree}, reinforcing its practicality as a plug-and-play alignment solution.

\section{Related Work}
\subsection{LLM Alignment and Preference Optimization}
LLM alignment methods have evolved from the RLHF paradigm towards more stable and cost-effective preference optimization. Early RLHF approaches~\cite{ouyang2022training} relied on expensive human feedback and unstable reinforcement learning, raising scalability and generalization concerns. To address this, Direct Preference Optimization (DPO)~\cite{rafailov2023direct} reframed preference learning as contrastive loss minimization, eliminating explicit reward modeling and RL sampling to improve training stability and efficiency.
Building on DPO, subsequent work like SimPO~\cite{meng2024simpo} simplified reward modeling using implicit log-probabilities, and multi-objective DPO~\cite{zhou2023beyond} explored multi-dimensional constraints for balanced alignment. Although DPO and its variants offer advantages in efficiency and robustness, they fundamentally provide sequence-level and post-hoc guidance during training. This limits fine-grained control and dynamic adaptation during complex inference-time generation, failing to meet the demand for immediate, granular alignment.

\subsection{Inference-time Scaling}
Inference-time scaling enhances LLM performance by utilizing additional compute during inference. 
Many related techniques~\cite{wei2022chain,wang2022self,yao2023tree} improve reasoning by guiding models through structured problem-solving.
As this field advances, preference control during inference has attracted attention. 
The most direct approach is the Best-of-N (BoN), which involves sampling $N$ outputs from a reference policy and then selecting the one with the highest reward according to a pre-defined reward model~\cite{touvron2023llama}.
TPO~\cite{li2025test} employs external reward models to iteratively correct outputs during inference. These methods guide output without parameter updates but face limitations in exploration efficiency and fine-grained control in complex spaces.

As a heuristic search algorithm, MCTS has surfaced in LLM inference-time optimization without parameter changes.
MCTSr~\cite{zhang2024accessing} combines MCTS with Self-Refine to improve mathematical reasoning. 
ReST-MCTS~\cite{zhang2024rest} integrates reward functions and trajectory sampling for unsupervised reasoning enhancement. 
However, existing MCTS applications focus primarily on mathematical or planning tasks, and a systematic exploration of inference-time alignment remains absent. Crucially, a key challenge remains: how to leverage MCTS with real-time guidance toward aligned responses during inference, particularly when strong models’ capabilities outstrip the expressiveness of supervised signals.

\subsection{Weak-to-Strong Generalization}
The rapid advancement of model capabilities has exacerbated the ``superalignment problem"~\cite{openai2023introducing}: ensuring models align with human intent even when supervision does not encompass their full capability space. 
W2SG~\cite{burns2023weak} systematically introduced this problem, showing stronger models can generalize beyond weak supervised signals across various tasks.
Subsequent research has broadened W2SG's application in alignment. WSPO~\cite{zhu2024weak} guides strong model fine-tuning by translating distribution differences from a weak aligned model into DPO-like reward signals. 
CBS~\cite{zhou2024weak} employs beam search to dynamically filter multiple candidate paths generated by strong models, selecting the paths deemed optimal by a weaker model's scoring as the final output.
While W2SG research is showing initial promise, building a W2SG mechanism that can be used for dynamic, inference-time alignment—without relying on additional human annotation or model updates—remains a critical challenge in current AI alignment research.

\section{Limitations and Future Work}

Although W2S-AlignTree has made significant progress in inference-time alignment, we recognize that the method still has room for improvement in several aspects, which will be the focus of our future research:
\begin{itemize}
    \item \textbf{Computational Overhead}: The introduction of MCTS inevitably increases the computational overhead during inference. While we have effectively controlled this overhead by utilizing a weak model, making it acceptable in most application scenarios, it remains a concern when pursuing extreme performance or handling tasks that are highly sensitive to latency.
    \item \textbf{Generality of the Scoring Function}: The DPO scoring function we adopt provides an effective proxy signal for alignment, but its generality may be limited by the quality and type of its training data. For alignment tasks that require complex, nuanced, or rare preferences, relying solely on a single DPO score may be insufficient to fully capture all human preferences.
    \item \textbf{MCTS Parameter Settings}: The performance of MCTS is related to the choice of hyperparameters. Although these parameters demonstrate robustness on specific tasks, their settings are somewhat task-specific, which introduces a certain exploration cost for the widespread application of the method.
    \item \textbf{Lack of Online Learning Capability}: The current framework is an offline alignment method that utilizes a pre-trained weak model for alignment during inference. It cannot learn and improve continuously through interaction with the environment like RLHF methods.
\end{itemize}

Based on the observations above, we propose several directions for future research to further enhance the W2S-AlignTree framework:
\begin{itemize}
    \item \textbf{Hybrid MCTS Strategies}: To further optimize computational efficiency, we plan to explore more intelligent MCTS strategies. A key direction is to design adaptive MCTS mechanisms, for example, dynamically adjusting the chunk length $L$ or number of candidates $K$ based on a node's confidence or the quality of generated content. This aims to maximize computational resource utilization while guaranteeing search quality.
    \item \textbf{More Complex Alignment Scoring}: To improve the robustness of the scoring function, future work will consider introducing multi-dimensional alignment scoring, for instance, simultaneously scoring for helpfulness, harmlessness, or creativity. We can also explore combining DPO scores with evaluations from other models (e.g., contrastive models) to obtain more comprehensive and detailed alignment signals.
    \item \textbf{Integration with Online Learning}: An important direction is to combine MCTS with lightweight online alignment methods. The high-quality alignment data generated by MCTS during the search process can serve as a valuable resource to fine-tune the strong LLM via SFT or DPO, enabling a synergy between inference-time alignment and continuous learning.
    \item \textbf{Multimodal Alignment}: Extending W2S-AlignTree to the multimodal domain holds great promise. By leveraging preference information, we can guide MCTS's exploration in the multimodal output space to generate content that better aligns with human preferences.
    \item \textbf{Interpretability and Theoretical Analysis}: We plan to conduct in-depth interpretability research on how MCTS paths and scores influence and enhance the alignment behavior of LLMs. This will help us improve our understanding of the entire alignment process and lay a foundation for developing more theoretically grounded and predictable alignment methods.
\end{itemize}

In summary, W2S-AlignTree is a promising inference-time alignment method that achieves significant advantages in alignment quality through MCTS search and the W2SG mechanism.
While it involves a certain trade-off in inference latency and parameter settings, we believe these limitations can be progressively overcome by incorporating more intelligent search strategies, more comprehensive scoring mechanisms, and integration with online learning. Future research will be dedicated to building a more efficient, robust, and scalable framework, with the aim of providing new perspectives and tools for the field of LLM alignment.

\end{document}